\documentclass[11pt]{article}

\usepackage[preprint]{acl}

\usepackage{times}
\usepackage{latexsym}
\usepackage{subcaption}
\usepackage{booktabs} % for professional tables
\usepackage{hyperref}
\usepackage{amsmath}
\usepackage{amssymb}
\usepackage{mathtools}
\usepackage{amsthm}
\usepackage{multirow}
\usepackage{makecell} % 单元格内换行
\usepackage{overpic}
\usepackage{xcolor}
\usepackage{listings}
\usepackage{tcolorbox}
\usepackage[table]{xcolor} % 用于显示灰色背景
\tcbuselibrary{listings, skins, breakable}
\usepackage{pifont}
\usepackage{graphbox}
\usepackage{fontawesome}
\usepackage{marvosym}
\usepackage{footmisc}
\definecolor{DeepGreen}{HTML}{228B22}
\definecolor{DeepRed}{HTML}{CF0000}

\newtcblisting{promptbox}[2][]{
    colback=gray!5,       % 背景色：淡灰
    colframe=gray!60!black, % 边框色：深灰
    arc=2pt, outer arc=2pt, % 圆角
    boxrule=0.8pt,        % 边框宽度
    listing only,         % 只显示内容，不执行代码
    listing options={
        basicstyle=\ttfamily\small, % 字体：等宽，小号
        breaklines=true,      % 自动换行
        breakatwhitespace=true, % 尽量在空格处换行
        columns=fullflexible, % 优化字符间距
        keepspaces=true,      % 保留空格缩进
        language=,            % 不强制语法高亮，适合纯文本/Markdown
        aboveskip=0pt,
        belowskip=0pt,
        showstringspaces=false
    },
    title={\textbf{#2}},  % 标题加粗
    #1
}

\definecolor{promptbg}{RGB}{248,248,248}
\definecolor{promptframe}{RGB}{210,210,210}
\definecolor{prompttitle}{RGB}{55,55,55}

\lstdefinestyle{promptstyle}{
    basicstyle=\ttfamily\small,
    breaklines=true,
    numbers=none,
    breakatwhitespace=false,
    columns=fullflexible,
    keepspaces=true,
    showstringspaces=false,
    tabsize=2,
    upquote=true
}

\newtcblisting{promptbox2}[2][]{
    enhanced,
    listing only,
    listing style=promptstyle,
    colback=promptbg,
    colframe=promptframe,
    coltitle=prompttitle,
    fonttitle=\bfseries\small,
    title={#2},
    boxrule=0.6pt,
    arc=2pt,
    left=6pt,
    right=6pt,
    top=6pt,
    bottom=6pt,
    before skip=6pt,
    after skip=6pt,
    width=\linewidth,
    #1
}

\definecolor{background}{HTML}{EEEEEE}
\definecolor{delim}{RGB}{112, 48, 160}
\definecolor{punct}{RGB}{80, 80, 80}
\colorlet{numb}{magenta!60!black}

\definecolor{exkeycolor}{RGB}{192, 79, 21}      % 外层 Key
\definecolor{inkeycolor}{RGB}{0, 112, 192}      % 内层 Key
\definecolor{background}{HTML}{F8F8F8}

\lstdefinelanguage{json}{
    basicstyle=\small\ttfamily,
    showstringspaces=false,
    breaklines=true,
    frame=none,
    xleftmargin=1em,     % 增加左侧缩进，补偿去掉行号后的视觉空白
    xrightmargin=1em,    % 左右对称设置，视觉更稳重
    moredelim=[is][\color{exkeycolor}]{\%}{\%}, 
    moredelim=[is][\color{inkeycolor}]{\#}{\#}, 
    backgroundcolor=\color{background},
    literate=
      {:}{{{\color{punct}{:}}}}{1}
      {,}{{{\color{punct}{,}}}}{1}
      {\{}{{{\color{delim}{\{}}}}{1}
      {\}}{{{\color{delim}{\}}}}}{1}
      {[}{{{\color{delim}{[}}}}{1}
      {]}{{{\color{delim}{]}}}}{1},
}

\usepackage[capitalize,noabbrev]{cleveref}

\usepackage[T1]{fontenc}
\usepackage[utf8]{inputenc}

\usepackage{microtype}

\usepackage{inconsolata}

\usepackage{graphicx}

\title{UniAudio-Token: Empowering Semantic Speech Tokenizers \\ with General Audio Perception}

\author{
    \textbf{Yuhan Song\textsuperscript{1}}\thanks{Work done during Yuhan's internship at WeChat AI.}\textmd{,}
    \textbf{Linhao Zhang\textsuperscript{2}\thanks{Corresponding authors.}}\textmd{,}
    \textbf{Aiwei Liu\textsuperscript{2}}\textmd{,}
    \textbf{Chuhan Wu\textsuperscript{2}}\textmd{,}
    \\ 
    \textbf{Sijun Zhang\textsuperscript{2}},
    \textbf{Wei Jia\textsuperscript{2}},
    \textbf{Yuan Liu\textsuperscript{2}},
    \textbf{Houfeng Wang\textsuperscript{1}\footnotemark[2]},
    \textbf{Xiao Zhou\textsuperscript{2}}
    \\
    \\
    \textbf{\textsuperscript{1}}State Key Laboratory of Multimedia Information Processing, \\School of Computer Science, Peking University
    \\
    \textbf{\textsuperscript{2}}Basic Model Technology Center, WeChat AI, Tencent Inc.
    \\
    \texttt{${\text{\small\Letter\;}}$\small{{\{songyuhan,wanghf\}@pku.edu.cn}\quad{zhanglinhao90@gmail.com}}}
}

\begin{document}
\maketitle
\begin{abstract}

% Semantic speech tokenizers have become the dominant foundation for Audio-LLMs. However, their exclusive focus on linguistic abstraction results in acoustic blindness, strictly limiting their utility to speech tasks while suppressing general audio cues. In this paper, we propose UniAudio-Token, a framework that empowers semantic tokenizers with general audio understanding without compromising their speech ability. Rather than altering the semantic paradigm, we rectify its information loss through two key innovations: (1) Semantic-Acoustic Primitives (SAP), a structured supervision protocol that decomposes raw audio into fundamental linguistic, paralinguistic, and environmental auditory scenes; and (2) Semantic-Acoustic Equilibrium (SAE), a content-aware gating mechanism that dynamically recovers fine-grained acoustic details from shallow layers. Extensive evaluations first validate the effectiveness of UniAudio-Token at the tokenizer level, demonstrating robust universal representation while preserving high-fidelity speech generation. When integrated with a lightweight 3B downstream LLM, it achieves an accuracy of 61.10\% on the MMAU benchmark, surpassing all the single-codebook baseline tokenizers.

Semantic speech tokenizers have become a widely used interface for Audio-LLMs, owing to their compact single-codebook design and strong linguistic alignment. However, their focus on linguistic abstraction induces \textit{acoustic blindness}, limiting their applicability beyond speech-centric tasks. We propose \textbf{UniAudio-Token}, a framework that empowers semantic tokenizers with general audio perception without compromising speech ability. Instead of altering the semantic paradigm, UniAudio-Token mitigates its information loss through two key innovations: (1) Semantic-Acoustic Primitives (SAP) provide structured supervision by decomposing audio into linguistic content, vocal attributes, and auditory-scene primitives; and (2) Semantic-Acoustic Equilibrium (SAE) introduces a content-aware gating mechanism that adaptively restores fine-grained acoustic details from shallow layers. Extensive evaluations show that UniAudio-Token learns comprehensive universal representations while preserving high-fidelity speech generation. When integrated with downstream LLMs, it outperforms all single-codebook baseline tokenizers on both understanding and generation tasks, effectively serving as a unified audio interface.
We publicly release all our code, including training and inference scripts, together with the model checkpoints at \url{https://github.com/Tencent/Universal_Audio_Tokenizer}.

\end{abstract}

\section{Introduction}
\label{sec:intro}

% Why discrete tokens instead of continuous features?
Audio-LLMs aim to extend LLMs to spoken and auditory interaction, requiring audio representations that support both \textbf{understanding} and \textbf{generation}. While continuous features from pretrained audio encoders are effective for perception, they significantly struggle with audio generation~\citep{yang-etal-2025-large-language}. In contrast, discrete audio tokens can be handled in the same modeling paradigm as text tokens, enabling a unified token-level interface for both audio input and output. This property has motivated recent Audio-LLMs to continue adopting discrete audio tokenizers~\cite{zeng2024glm,ding2025kimi,zhang2025mimo}, and makes improving audio tokenizers a critical problem rather than merely an architectural choice.

% Paragraph 1: Background - Semantic Tokenizers are Widely Adopted
Among discrete audio tokenizers, \textbf{semantic speech tokenizers} have been widely adopted in recent Audio-LLMs~\cite{du2024cosyvoice2,zeng2025scaling,song2025stabletoken} due to two compelling advantages: (1) single-codebook design, which enables direct integration into standard LLM architectures and compact sequences crucial for long-context processing; and (2) inherent linguistic alignment, as initialization from ASR encoders facilitates seamless text-audio interaction.

\begin{table}[t]
    \centering
\resizebox{\linewidth}{!}{
    \begin{tabular}{lccc}
        \toprule
        \multirow{2}{*}{\textbf{Model}} & \textbf{Single} & \textbf{General} & \textbf{Linguistic} \\
        & \textbf{Codebook} & \textbf{Audio} & \textbf{Alignment} \\
        \midrule
        EnCodec~\cite{defossez2023high} & \color{DeepRed}{\ding{55}} & \color{DeepGreen}{\ding{51}} & \color{DeepRed}{\ding{55}} \\
        SpeechTokenizer~\cite{zhang2024speechtokenizer} & \color{DeepRed}{\ding{55}} & \color{DeepGreen}{\ding{51}} & \color{DeepGreen}{\ding{51}} \\
        CosyVoice2~\cite{du2024cosyvoice2} & \color{DeepGreen}{\ding{51}} & \color{DeepRed}{\ding{55}} & \color{DeepGreen}{\ding{51}} \\
        GLM-4-Voice-Tokenizer~\cite{zeng2025scaling} & \color{DeepGreen}{\ding{51}} & \color{DeepRed}{\ding{55}} & \color{DeepGreen}{\ding{51}} \\
        StableToken~\cite{song2025stabletoken} & \color{DeepGreen}{\ding{51}} & \color{DeepRed}{\ding{55}} & \color{DeepGreen}{\ding{51}} \\
        WavTokenizer~\cite{ji2025wavtokenizer} & {\color{DeepGreen}{\ding{51}}} & \color{DeepGreen}{\ding{51}} & \color{DeepRed}{\ding{55}} \\
        \rowcolor{gray!10} \textbf{UniAudio-Token (Ours)} & \color{DeepGreen}{\ding{51}} & \textbf{\color{DeepGreen}{\ding{51}}} & \textbf{\color{DeepGreen}{\ding{51}}} \\
        \bottomrule
    \end{tabular}
}
    % \caption{Comparison of audio tokenizers. While semantic-centric models suffer from \textit{Acoustic Blindness} and acoustic-centric models suffer from \textit{Entanglement}, \textbf{UniAudio-Token} uniquely empowers semantic tokenizers with universal acoustic perception.}
    \caption{Comparison of audio tokenizers. UniAudio-Token uniquely combines single-codebook modeling, general audio perception, and linguistic alignment.}
    \label{tab:comparison}
\end{table}

% Paragraph 2: The Gap - Semantic Models suffer from "Acoustic Blindness"
However, as Audio-LLMs expand from speech to universal auditory perception, including music, sound events, and complex acoustic scenes, semantic speech tokenizers lag behind. Optimized strictly for linguistic content extraction, deep ASR encoders actively suppress vocal cues and auditory scene details as \textit{noise}. This induces \textbf{acoustic blindness}, fundamentally limiting the LLM's understanding of the full acoustic scene.

\begin{figure*}[tb]
    \centering
    \begin{minipage}[b]{0.2480975428\textwidth}
        \centering
        \includegraphics[width=\linewidth]{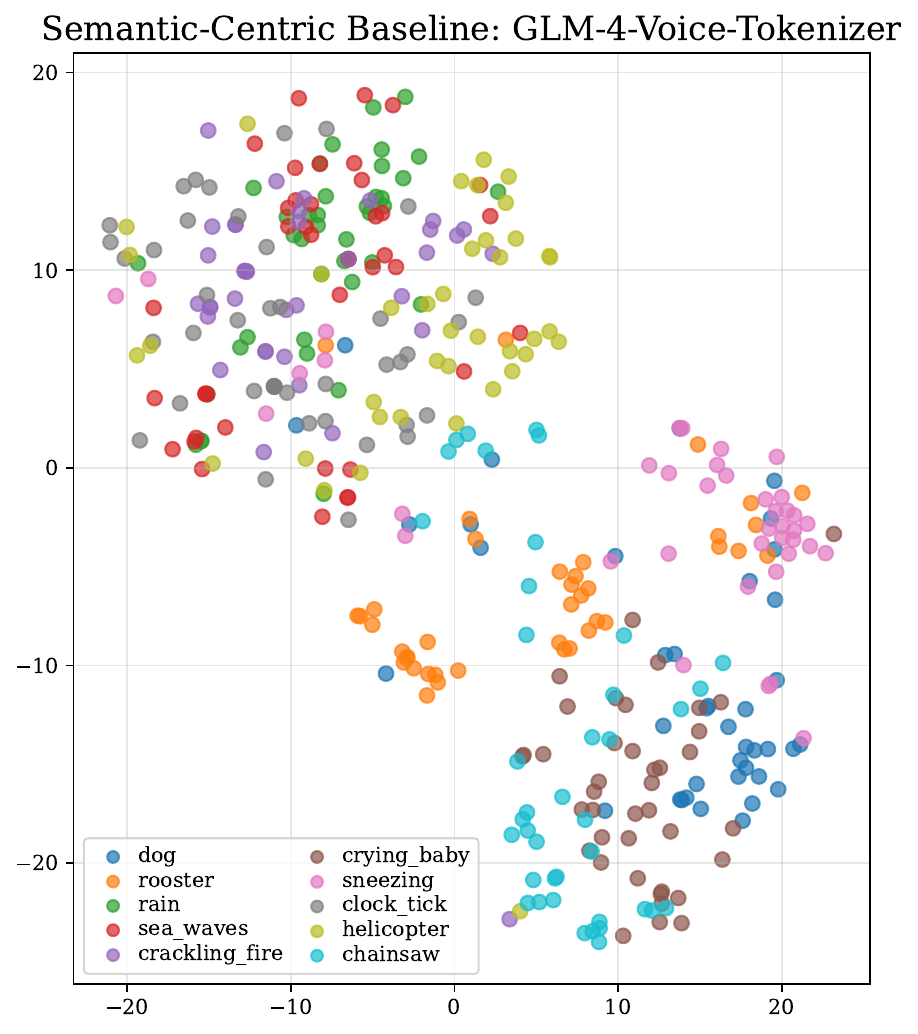}
    \end{minipage}
    \begin{minipage}[b]{0.2397012286\textwidth}
        \centering
        \includegraphics[width=\linewidth]{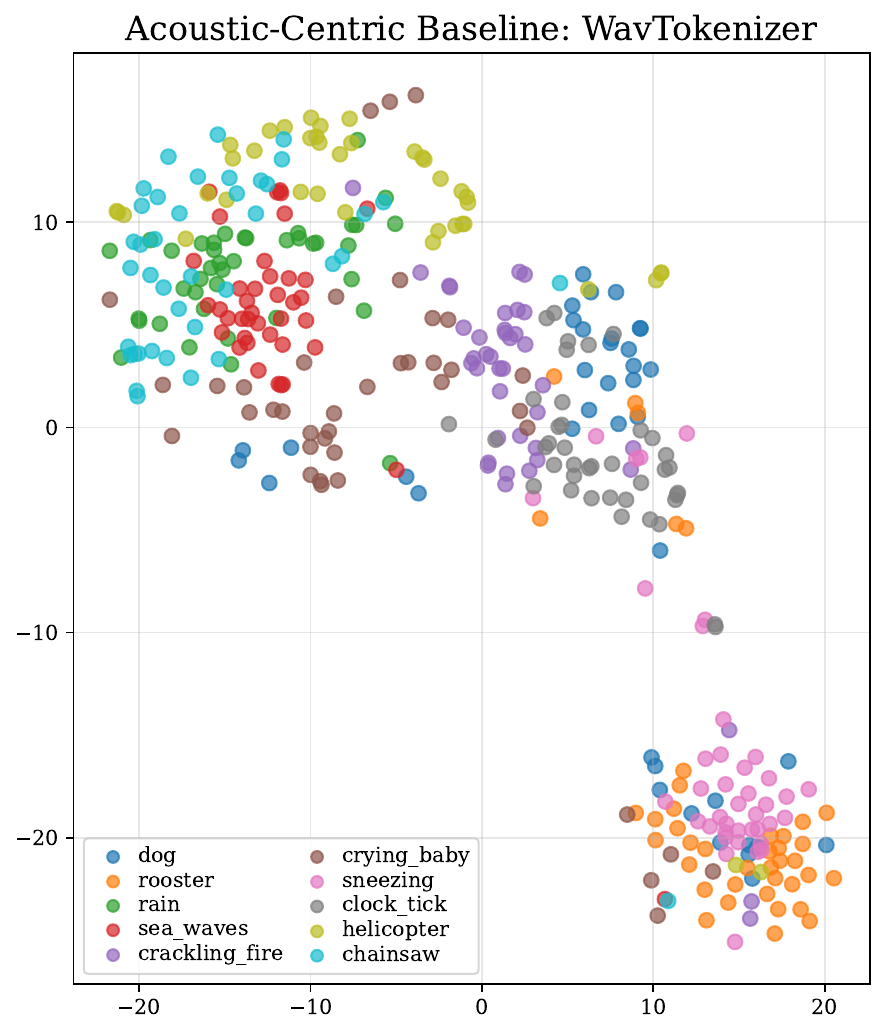}
    \end{minipage}
    \begin{minipage}[b]{0.2397012286\textwidth}
        \centering
        \includegraphics[width=\linewidth]{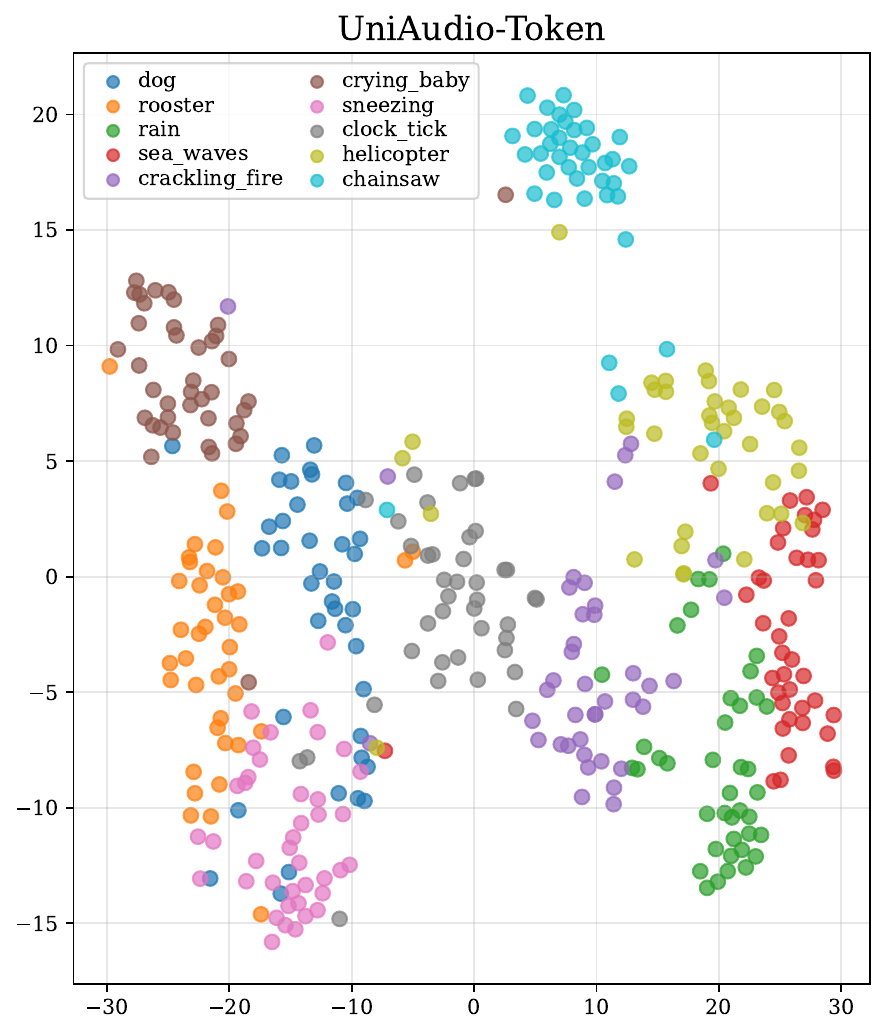}
    \end{minipage}
    {\color{gray!40}\vrule width 0.5pt}
    \begin{minipage}[b]{0.2425\textwidth}
        \centering
        \includegraphics[width=\linewidth]{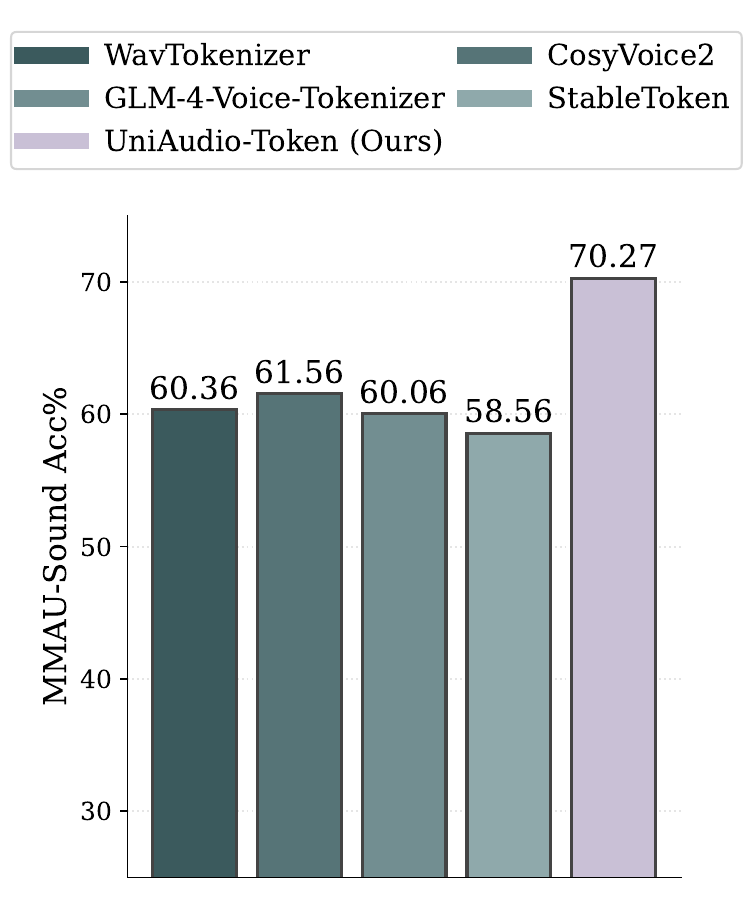}
    \end{minipage}
    \caption{ESC-10 token sequence t-SNE Visualization. \textit{(Left)} A semantic-centric baseline (GLM-4-Voice-Tokenizer) suffers from acoustic blindness, mapping distinct events to overlapping regions. \textit{(Center Left)} An acoustic-centric baseline (WavTokenizer) exhibits insufficient semantic discrimination. \textit{(Center Right)} UniAudio-Token resolves these issues via Semantic-Acoustic Equilibrium, forming well-separated clusters. \textit{(Right)} When integrated with Qwen2.5-3B, UniAudio-Token shows superior performance on the MMAU benchmark.}
    % \caption{ESC-10 token t-SNE and MMAU results. GLM-4-Voice suffers from acoustic blindness, WavTokenizer lacks semantic discrimination, while UniAudio-Token forms well-separated clusters and leads ahead on MMAU.}
    \label{fig:teaser}
\end{figure*}

% Paragraph 3: Alternative Approaches - The Trade-off
Alternatively, single-codebook acoustic-centric models~\cite{ji2025wavtokenizer} prioritize waveform reconstruction. While preserving acoustic details, they lack explicit semantic guidance. Consequently, their audio tokens fail to form distinct categorical clusters based on meaning. Semantically distinct but acoustically similar sounds (e.g. \textit{rain} and \textit{white noise}) may collapse into overlapping distributions.
% Consequently, their audio tokens fail to form distinct categorical clusters based on meaning. Semantically distinct sounds with similar acoustic textures (e.g., \textit{rain} and \textit{white noise}) are conflated into overlapping distributions. 
% This absence of semantic separability imposes a heavy burden on the LLM, creating a harsh trade-off between capturing acoustic details and extracting linguistic meaning.

% This limitation has forced a fragmented paradigm in current Audio-LLM research: models are either confined to speech-only understanding \cite{zeng2024glm}, or they must resort to a heterogeneous architecture—relying on an external continuous encoder via adapters for general audio perception, while reserving the tokenizer for speech generation \cite{huang2025step,Qwen2.5-Omni}.

This harsh semantic-acoustic trade-off forces a fragmented paradigm: Audio-LLMs are either confined to speech-only understanding~\cite{zeng2024glm}, or rely on heterogeneous architectures that combine external continuous encoders and adapters for general audio perception with discrete tokenizers for speech generation~\cite{huang2025step,Qwen2.5-Omni}. In this work, we aim to unify this divide by using a single codebook to support both high-fidelity \textbf{speech generation} and high-level \textbf{general audio understanding} simultaneously.

% % Paragraph 4: The Challenge
% Empowering semantic tokenizers with universal audio understanding is non-trivial and involves two fundamental conflicts: (1) \textit{The supervision conflict}: ASR targets~\cite{zeng2024glm,du2024cosyvoice2,song2025stabletoken} extract linguistics but ignore acoustic details, whereas reconstruction targets~\cite{ji2025wavtokenizer} excessively focus on raw acoustic nuances, hindering clean semantic extraction. Existing semantic distillation approaches~\cite{zhang2024speechtokenizer}, often based on teachers like HuBERT~\cite{hsu2021hubert} or Whisper~\cite{radford2023robust}, remain predominantly speech-centric and fail to generalize to general audio. (2) \textit{The architectural bottleneck}: Deep semantic encoders irreversibly lose fine-grained acoustic cues in higher layers, while naive feature fusion risks diluting the linguistic abstraction required for content-faithful speech generation.
% A mechanism that dynamically balances these competing information streams is required.

% Paragraph 4: The Challenge
Empowering semantic tokenizers with universal audio perception is non-trivial. It involves two fundamental conflicts: (1) \textit{The supervision conflict}: ASR targets~\cite{du2024cosyvoice2,zeng2025scaling,song2025stabletoken} extract linguistics but ignore acoustics, whereas reconstruction targets~\cite{ji2025wavtokenizer} focus on raw acoustic nuances, hindering semantic extraction. Semantic distillation~\cite{zhang2024speechtokenizer}
%, often based on teachers like HuBERT~\cite{hsu2021hubert} or Whisper~\cite{radford2023robust}, 
remains speech-centric and fails to generalize to general audio. 
Recent work has begun exploring supervision beyond transcription~\citep{zhang2026transcriptionunifiedaudioschema}, but only at the Audio-LLM level, without addressing acoustic blindness at the audio tokenizer level, which is the fundamental representational bottleneck. (2) \textit{The architectural bottleneck}: Deep semantic encoders irreversibly lose fine-grained acoustic cues in higher layers, while naive feature fusion risks diluting the linguistic abstraction required for content-faithful speech generation.
A mechanism that dynamically balances these competing information streams is required.

% Paragraph 5: Our Solution
To address these challenges, we propose the \textbf{UniAudio-Token} framework to empower single-codebook semantic speech tokenizers with universal audio perception. Our core insight is that mitigating the semantic-acoustic tension requires dual rectification: explicitly disentangling linguistic content from vocal attributes and auditory scenes \textit{at the supervision level}, and dynamically bridging the information bottleneck to recover lost acoustic details \textit{at the architectural level}.

Specifically, we introduce two innovations:
(1) \textbf{Semantic-Acoustic Primitives (SAP):} Resolving the supervision conflict, this structured supervision protocol decomposes raw audio into fundamental linguistic content, vocal attributes, and auditory-scene building blocks. It explicitly disentangles content from style, forcing the model to allocate capacity for vocal and acoustic details without interfering with the semantic backbone.
(2) \textbf{Semantic-Acoustic Equilibrium (SAE):} Addressing the architectural bottleneck, this content-aware gating mechanism adaptively injects fine-grained acoustic details from shallow layers into deep semantic streams when needed, mitigating acoustic blindness without corrupting semantic representations.

% Paragraph 6: Results & Disclaimer
Extensive evaluations demonstrate UniAudio-Token effectively bridges linguistic alignment and universal representation. At the tokenizer level, it achieves high Cluster Purity on ESC~\citep{piczak2015esc}, forming distinct clusters for diverse audio events where baselines struggle. Crucially, this acquisition of general audio perception does not compromise speech generation capabilities; instead, UniAudio-Token even surpasses specialized speech tokenizers in generation quality. At the Audio-LLM level, integrating this universal frontend with Qwen2.5 also yields superior performance on both understanding and generation. Further analysis validates the adaptive behavior of the SAE mechanism.

\section{Related Work}
\label{sec:related_work}

\paragraph{Semantic Speech Tokenizers.}
The evolution of LLMs has pushed spoken dialogue systems from traditional cascaded pipelines towards end-to-end Audio-Language Models ~\citep{zhang2019using,zhang2020graph,tang2023salmonn,zhang2023speechgptempoweringlargelanguage,gong2024listenthinkunderstand,hu2024wavllm,fang2024llama,defossez2024moshi,li2025baichuan,wang2024freeze,bai2024audiosetcapsenrichedaudiocaptiondataset,goel2025audioflamingo3advancing,zhang2025wildspeech}, driving two distinct tokenizer paradigms.
Early self-supervised learning (SSL) units~\citep{hsu2021hubert,baevski2020wav2vec,huang2022spiral}
% While effective for discriminative tasks, these tokens 
primarily encode phonetic information and suffer from high Gross Pitch Error in generation~\citep{sicherman2023analysing,mousavi2024dasb}, making them unsuitable for high-fidelity end-to-end synthesis.
Recent systems adopt supervised ASR-based tokenization~\citep{du2024cosyvoice2,ding2025kimi,song2025stabletoken}, quantizing intermediate representations of ASR encoders into compact flat tokens that represent linguistic units while implicitly retaining prosodic features learned from large-scale transcription.
However, despite their success in speech understanding and synthesis, we find this paradigm fundamentally suffers from \textit{acoustic blindness} in general audio tasks. 

\paragraph{Acoustic Audio Tokenizers.}
In parallel, acoustic tokenizers target high-fidelity waveform reconstruction. Neural codecs~\citep{zeghidour2021soundstream,defossez2023high,yang2023hifi,kumar2023high} typically employ multi-codebook RVQ to reduce distortion, which necessitates specialized architectural adaptations or flattening for LLM integration. 
% However, integrating such structures with standard LLMs requires specific architectural adaptations or flattening strategies, lacking the seamless, out-of-the-box integration characteristic of single-codebook designs.
Recent acoustic-centric models like WavTokenizer~\citep{ji2025wavtokenizer} instead employ single-codebook vector quantization to achieve extreme compression with a flat structure. While this facilitates direct LLM integration and preserves general audio details, it lacks explicit semantic alignment, limiting the performance on linguistic-intensive tasks. Although semantic distillation techniques~\citep{defossez2024moshi,zhang2025mimo,ye2025codec} can improve semantic awareness, they retain multi-codebook design and restrict semantic supervision to speech, leaving general audio events semantically entangled.
% hinton2015distillingknowledgeneuralnetwork, cai-etal-2022-pile,

\section{Methods}
\label{sec:methods}

\begin{figure*}[tb]
    \centering
    \includegraphics[width=0.85\textwidth]{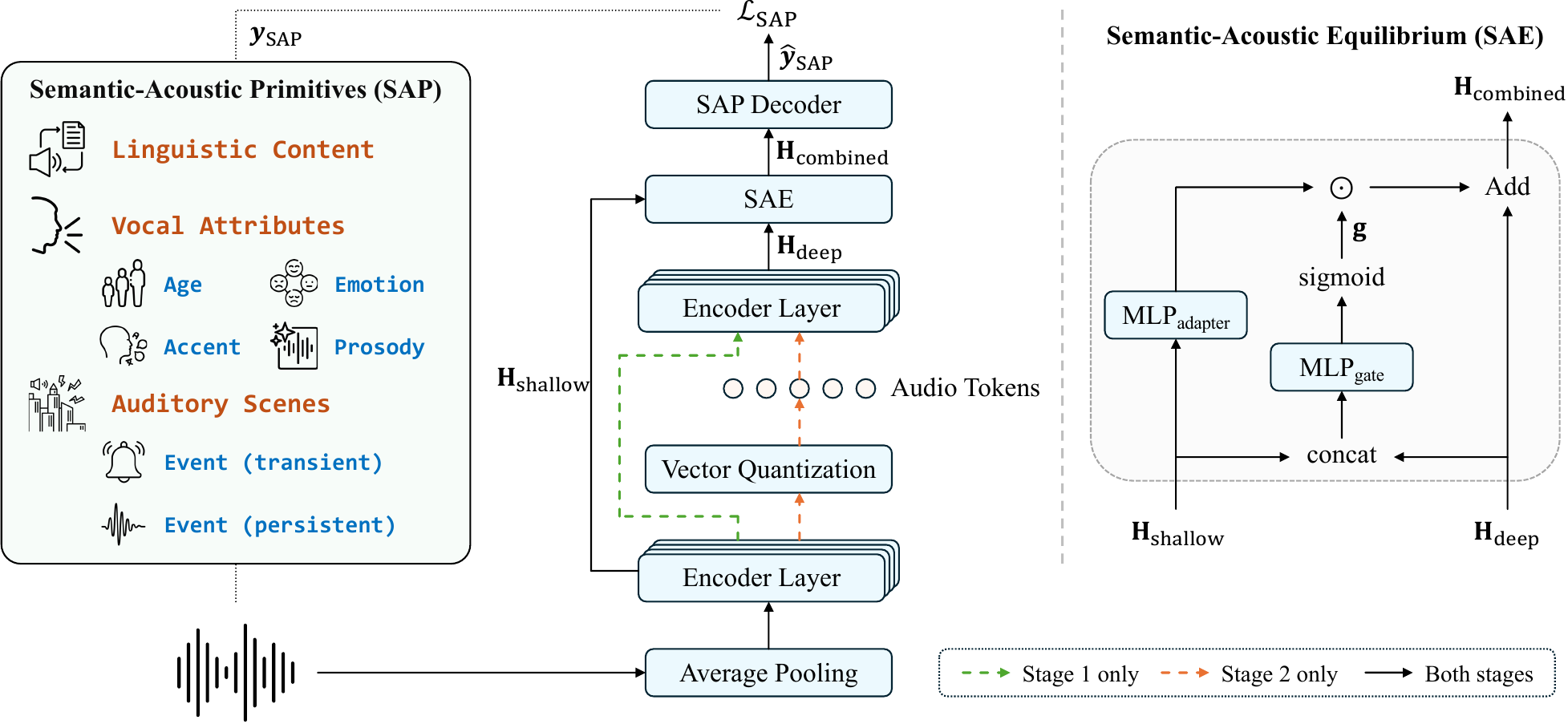}
    \caption{The framework of UniAudio-Token. \textit{(Left)} The model is supervised by Semantic-Acoustic Primitives (SAP), which cover linguistic content, vocal attributes, and auditory scenes.  \textit{(Center)} Vector Quantization (VQ) converts hidden states into discrete audio tokens. \textit{(Right)} Semantic-Acoustic Equilibrium (SAE) adaptively fuses shallow acoustic details with deep semantic features, mitigating the loss of fine-grained acoustic cues in deep layers.}
    \label{fig:model}
\end{figure*}

To resolve the architectural fragmentation of Audio-LLMs, \textbf{UniAudio-Token} establishes a \textit{unified discrete interface} that projects speech and general audio into a single codebook. As shown in Figure~\ref{fig:model}, we reconcile linguistic generation and universal perception via two core innovations: Semantic-Acoustic Primitives (SAP) for disentangled supervision, and Semantic-Acoustic Equilibrium (SAE) for adaptive semantic-acoustic feature fusion.

\subsection{Semantic-Acoustic Primitives (SAP)}
\label{sec:mas_data}

Existing audio tokenizers face a fundamental trade-off: ASR-based supervision provides strong linguistic alignment but limited discrimination for non-speech signals, while reconstruction-based objectives preserve acoustic details but lack explicit semantic guidance. To address this limitation, we introduce a structured supervision strategy termed \textbf{Semantic-Acoustic Primitives (SAP)}. SAP serves as a protocol generated by an LLM to provide supervision that captures both semantic content and acoustic cues. Unlike traditional ASR corpora that focus solely on linguistic content, SAP explicitly separates and annotates the full spectrum of acoustic information, enabling the tokenizer to remain discriminative across diverse audio types.

% \subsubsection{MAS Structure}
\paragraph{Structure Design.}
SAP describes each audio clip using three complementary layers: (1) \textbf{Linguistic Content}, i.e., the verbatim transcript for speech; (2) \textbf{Vocal Attributes}, which characterize \textit{how} speech is produced through six normalized fields: \textit{Age, Gender, Emotion, Accent, Prosody,} and \textit{Timbre}; and (3) \textbf{Auditory Scene}, which captures the acoustic environment, including \textit{Transient Events} (e.g., door slams) and \textit{Persistent Events} (e.g., engine rumble).

% This structured format provides low-entropy supervision targets, stabilizing the optimization of the tokenizer by explicitly separating semantic meaning from acoustic texture. By providing disentangled gradients, SAP constrains the codebook to allocate capacity for both linguistic and non-linguistic information.
This structure separates semantic meaning from acoustic cues explicitly and provides low-entropy supervision targets, thereby stabilizing tokenizer optimization. Example JSON annotations are provided in Appendix~\ref{app:sap_sample} for better understanding.

% \subsubsection{Scalable Generation Pipeline}
\paragraph{Data Curation.}
Since manually annotating such fine-grained attributes is prohibitively costly, we develop an automated pipeline to derive SAP labels from large-scale ASR corpora: (1) \textbf{Acoustic Captioning.} An audio-language model generates rich, unstructured textual descriptions of the audio, capturing vocal style and auditory-scene information missing from ASR transcripts. (2) \textbf{Structured Synthesis.} An LLM teacher aggregates the ground-truth transcription and the generated acoustic captions, normalizes them into predefined SAP fields, and outputs a valid JSON object. (3) \textbf{Quality Validation.} We apply a multi-level validation mechanism to reduce hallucinations, including ontology constraints for categorical fields, logical consistency checks, and content-duration alignment. Only samples passing all checks are retained. Human evaluation further verifies the reliability of the SAP annotations, with details in Appendix~\ref{app:human_eval}.

% \subsubsection{Instruction Tuning Data (SAP-Instruct)}
% To facilitate interactive capabilities, we further derive a \textbf{SAP-Instruct} dataset from the structured annotations. This includes Direct QA (e.g., ``What is the speaker's emotion?''), Multiple Choice, and Binary Verification (Yes/No) pairs. This diverse format forces the model to attend to specific acoustic sub-features during training.
To facilitate interactive capabilities, we further derive an \textbf{SAP-Instruct} dataset from the structured annotations, including Direct QA, Multiple Choice, and True/False Verification pairs. These diverse formats encourage the model to attend to specific acoustic sub-features during training.

\subsection{Model Architecture}
As illustrated in Figure~\ref{fig:model}, UniAudio-Token consists of an audio encoder, an SAE module, a quantization layer, and an SAP decoder.

% Unlike previous semantic tokenizers \cite{zeng2024glm,song2025stabletoken,du2024cosyvoice} that focus solely on ASR transcription, our model is supervised by SAP, which encompass linguistic content, vocal features and auditory scenes. This requires the discrete tokens to encode much richer information than standard semantic tokens.

\paragraph{Semantic-Acoustic Equilibrium (SAE).}
ASR-centric speech encoders, such as Whisper~\citep{radford2023robust}, progressively abstract audio into high-level semantic representations. While beneficial for ASR, this process often discards low-level acoustic details required by SAP, such as vocal texture and auditory events. The \textbf{Semantic-Acoustic Equilibrium (SAE)} mechanism addresses this bottleneck by adaptively fusing semantically rich deep features with acoustically rich shallow features.

Let $\mathbf{H}_{\text{shallow}}$ denote the output from a shallow encoder layer, and $\mathbf{H}_{\text{deep}}$ denote the final-layer representation. First, we project the shallow features into the deep feature space: 
\begin{equation}
    % \small
    \mathbf{H}_{\text{ada\_shallow}} = \mathbf{MLP}_{\text{adapter}}\left(\mathbf{H}_{\text{shallow}}\right),
\end{equation}
where $\mathbf{MLP}_{\text{adapter}}$ is learnable. Then the SAE computes a content-aware fusion gate $\mathbf{g}$:
\begin{equation}
    % \small
    \mathbf{g} = \sigma\left(\mathbf{MLP}_{\text{gate}} \left([\mathbf{H}_{\text{deep}} ; \mathbf{H}_{\text{shallow}}]\right)\right),
\end{equation}
where $[\cdot;\cdot]$ denotes concatenation and $\sigma$ is the sigmoid function. $\mathbf{MLP}_{\text{gate}}$ is learnable. The final fused representation $\mathbf{H}_{\text{combined}}$ is then obtained via:
\begin{equation}
    % \small
    \mathbf{H}_{\text{combined}} = \mathbf{H}_{\text{deep}} + \mathbf{g} \odot \mathbf{H}_{\text{ada\_shallow}},
\end{equation}
where $\odot$ denotes element-wise multiplication. The SAE mechanism allows the model to adaptively retain acoustic details necessary for SAP-supervised learning while preserving semantic abstraction.

\paragraph{Vector Quantization.}
We discretize the continuous hidden states using a standard Vector Quantization (VQ) layer~\citep{van2017neural}. Given a learnable codebook $\mathcal{C} = \{\mathbf{e}_k\}_{k=1}^{K} \subset \mathbb{R}^D$, where $K$ is the codebook size, the input vector $\mathbf{h}_t$ at time step $t$ is mapped to its nearest code vector:
\begin{equation}
    % \small
    \mathbf{h}_{t}^{q} = \mathbf{e}_k, \quad \text{where } k = \mathop{\text{argmin}}_{j} \|\mathbf{h}_t - \mathbf{e}_j\|_2^2.
\end{equation}
The sequence of indices $k$ forms the audio tokens.

\subsection{Training Strategy}
\label{sec:training}
We initialize the encoder and decoder from whisper-large-v3~\citep{radford2023robust} and train the full model end-to-end with a mixture of SAP generation and SAP-Instruct QA tasks. This initialization preserves strong linguistic alignment, while SAP supervision then expands the representation toward vocal attributes and auditory scenes.

Our training pipeline consists of two stages. In \textbf{Stage 1}, we bypass the VQ layer and train the SAE module together with the decoder using only SAP prediction loss ($\mathcal{L}_{\text{SAP}}$). The goal is to adapt the pretrained ASR decoder into an SAP decoder, aligning the continuous hidden space with structured SAP. In \textbf{Stage 2}, we insert the VQ layer and primarily optimize the codebook to produce discrete audio tokens, while preserving the SAP-aligned representation learned in the previous stage. Following the framework of VQ~\citep{van2017neural}, the objective function combines the SAP prediction loss with quantization and commitment losses:
\begin{equation}
    \small
    \mathcal{L} = \mathcal{L}_{\text{SAP}} + \lambda_1 \underbrace{\|\text{sg}[\mathbf{h}] - \mathbf{h}^q\|_2^2}_{\mathcal{L}_{\text{quantization}}} + \lambda_2 \underbrace{\| \mathbf{h} - \text{sg}[\mathbf{h}^q] \|^2_2}_{\mathcal{L}_{\text{commitment}}},
\end{equation}
where $\lambda_1, \lambda_2$ are hyperparameters, and $\text{sg}[\cdot]$ the stop-gradient operator. The decoder is optimized by $\mathcal{L}_{\text{SAP}}$, the encoder and SAE module by $\mathcal{L}_{\text{SAP}}$ and $\mathcal{L}_{\text{commitment}}$, and the codebook by $\mathcal{L}_{\text{quantization}}$.

% \paragraph{Data Mixture.}
% We employ a balanced mixture of 30\% SAP generation data and 70\% SAP-Instruct data during training. The SAP generation task requires the model to produce the complete structured annotation from audio, ensuring comprehensive acoustic encoding. The SAP-Instruct tasks (Direct QA, Multiple Choice, Binary Verification) force the model to attend to specific acoustic sub-features, preventing the codebook from collapsing to a narrow subset of acoustic attributes. This mixture strategy yields a universal audio tokenizer that maintains both holistic understanding and fine-grained discriminability.

\section{Experimental Setup}

% We utilize the Qwen2.5-3B model as the backbone LLM for understanding tasks. For the tokenizer, we compare UniAudio-Token against several state-of-the-art baselines, including \textbf{EnCodec} (32kbps), \textbf{WavTokenizer}, and \textbf{StableToken}. All models are trained on the same 50k-hour dataset consisting of speech, music, and environmental sounds.

\label{sec:setup}

% We structure the evaluation into four dimensions: (1) \textbf{Audio Understanding}, assessing the model's capability on audio understanding benchmarks when integrated with an LLM; (2) \textbf{Audio Reconstruction}, verifying the fidelity and quality of the discrete token representation; (3) \textbf{Visualization of Acoustic Event Separation}, a diagnostic analysis probing the distinctiveness of our token space; and (4) \textbf{Ablation Studies}, isolating the contribution of the SAE module.

\paragraph{Implementation Details.}
For SAP data curation, we utilize {Qwen3-Omni-Captioner}~\citep{Qwen3-Omni} to produce detailed acoustic captions and {Qwen3-30B-A3B-Instruct-2507}~\citep{yang2025qwen3} to perform structured synthesis. SAP-Instruct is curated using the more powerful {Qwen3-235B-A22B-Instruct-2507}~\citep{yang2025qwen3} to ensure high-quality instruction following.

UniAudio-Token uses a single codebook with a vocabulary size of 8,192 and a token frame rate of 25Hz. Full training details, including datasets and hyperparameters, are listed in Appendix~\ref{app:training_details}.

For downstream Audio-LLMs, we integrate all audio tokenizers with the same Qwen2.5~\citep{yang2024qwen2.5} LLM backbone and train them under the same settings for fair comparison.

\paragraph{Baselines.}
% We compare UniAudio-Token against several single-codebook neural audio codecs and semantic tokenizers, including: (1) \textbf{WavTokenizer} (Large, 75Hz) ~\citep{ji2025wavtokenizer}, representing high-compression acoustic codecs; (2) \textbf{CosyVoice2}~\citep{du2024cosyvoice2}, a leading speech tokenization and generation model; (3) \textbf{GLM-4-Voice-Tokenizer}~\citep{zeng2025scaling}, a representative semantic tokenizer tailored for speech LLMs; and (4) \textbf{StableToken}~\citep{song2025stabletoken}, a novel speech tokenizer with superior noise robustness. A detailed introduction to these baselines is provided in Appendix~\ref{app:baselines}.
We compare UniAudio-Token against representative single-codebook acoustic and semantic audio tokenizers, including (1) \textbf{WavTokenizer} (Large, 75Hz)~\citep{ji2025wavtokenizer}; (2) \textbf{CosyVoice2}~\citep{du2024cosyvoice2}; (3) \textbf{GLM-4-Voice-Tokenizer}~\citep{zeng2025scaling}; and (4) \textbf{StableToken}~\citep{song2025stabletoken}. An additional brief introduction of these baselines is provided in Appendix~\ref{app:baselines}.

\paragraph{Evaluation \& Benchmarks.}

We use t-SNE visualization~\citep{maaten2008visualizing} and clustering metrics (Silhouette Score~\citep{rousseeuw1987silhouettes} and Cluster Purity~\citep{manning2008introduction}) on ESC~\citep{piczak2015esc} to measure UniAudio-Token's discriminability across diverse sound events. We evaluate speech reconstruction and text-to-speech (TTS) synthesis on LibriSpeech~\citep{panayotov2015librispeech} and SEED-TTS~\citep{anastassiou2024seed} using content faithfulness (WER) and speech quality (MOS predicted by MOSNet~\citep{lo2019mosnet}).

We evaluate the tokenizer-LLM systems on MMAU~\citep{sakshi2025mmau}, MMAR~\citep{ma2025mmar}, and MMSU~\citep{wang2025mmsu}.
These comprehensive audio understanding tasks cover diverse audio inputs, including speech, music, and sounds.
% with performance measured via content faithfulness (WER), speech quality (MOS), speaker similarity (SIM), and Fr\'echet Audio Distance (FAD).

% \textbf{Analysis \& Ablation:} We employ t-SNE visualization for qualitative token analysis and standard metric drops for ablation studies.

\section{Results}
\label{sec:results}

\begin{figure*}[t]
    \centering
    % 左图：UniAudio，右图：GLM-4
    \begin{subfigure}{0.326666667\textwidth}
        \centering
        \includegraphics[width=\linewidth]{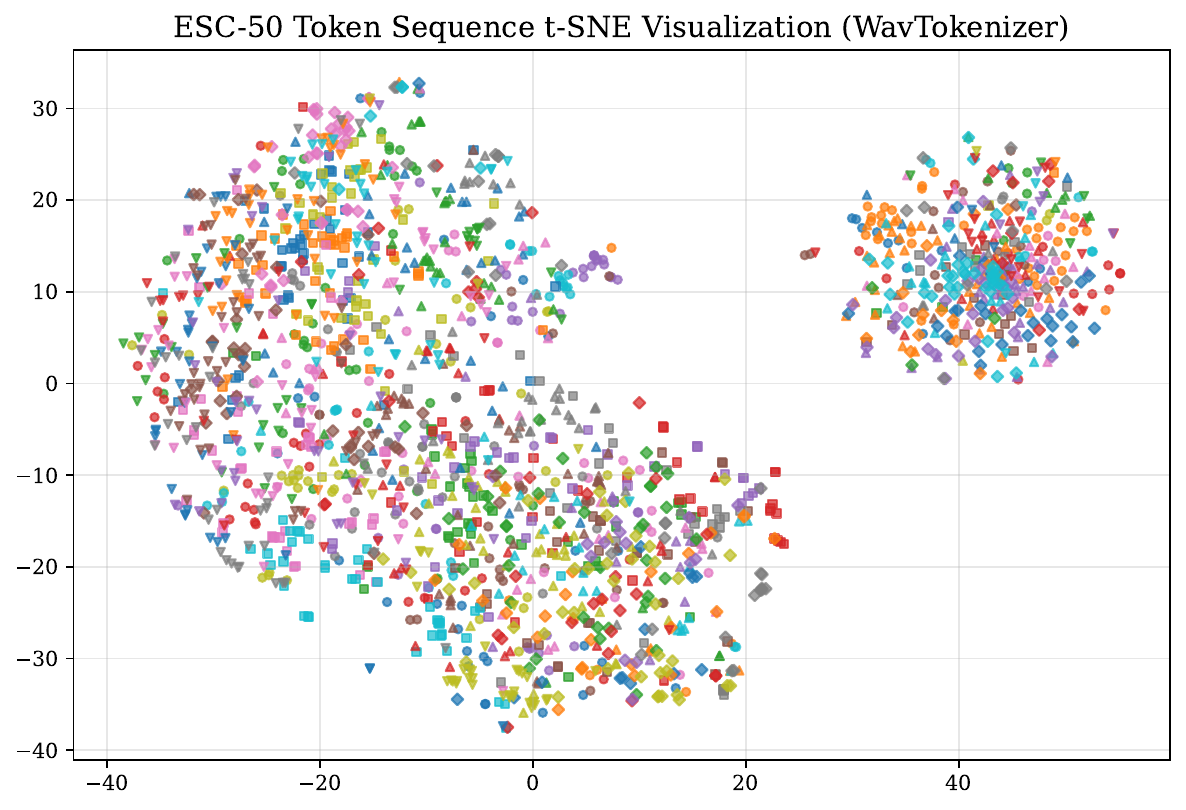}
        \caption{WavTokenizer}
        \label{fig:tsne_wav}
    \end{subfigure}
    \hfill
    \begin{subfigure}{0.326666667\textwidth}
        \centering
        \includegraphics[width=\linewidth]{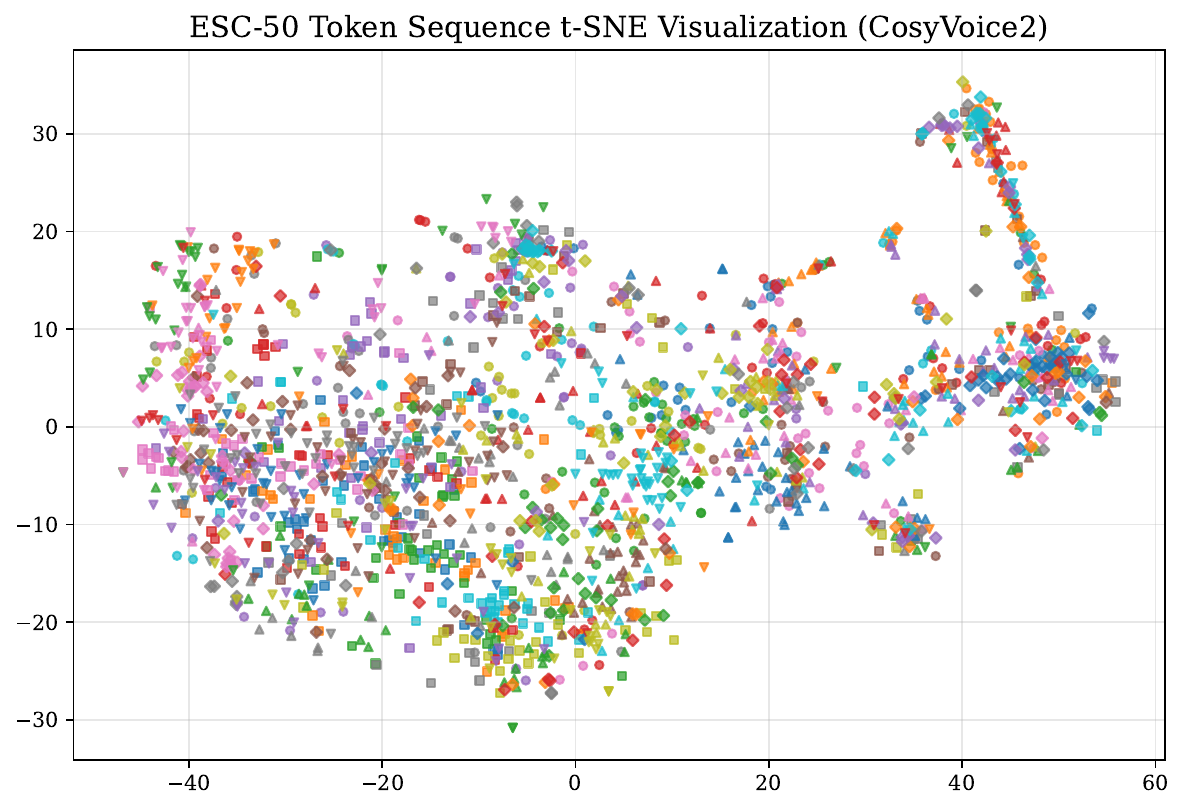}
        \caption{CosyVoice2}
        \label{fig:tsne_cos}
    \end{subfigure}
    \hfill
    \begin{subfigure}{0.326666667\textwidth}
        \centering
        \includegraphics[width=\linewidth]{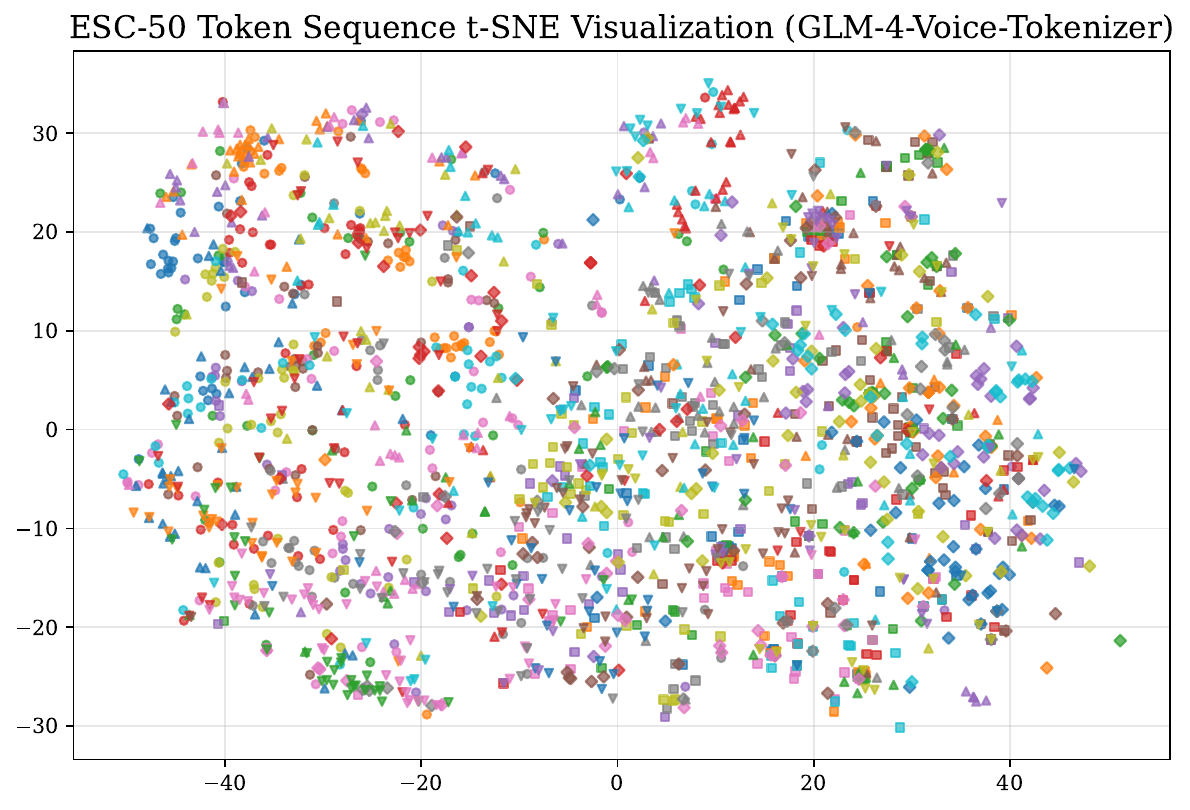}
        \caption{GLM-4-Voice-Tokenizer}
        \label{fig:tsne_glm}
    \end{subfigure}

    \begin{subfigure}{0.326666667\textwidth}
        \centering
        \includegraphics[width=\linewidth]{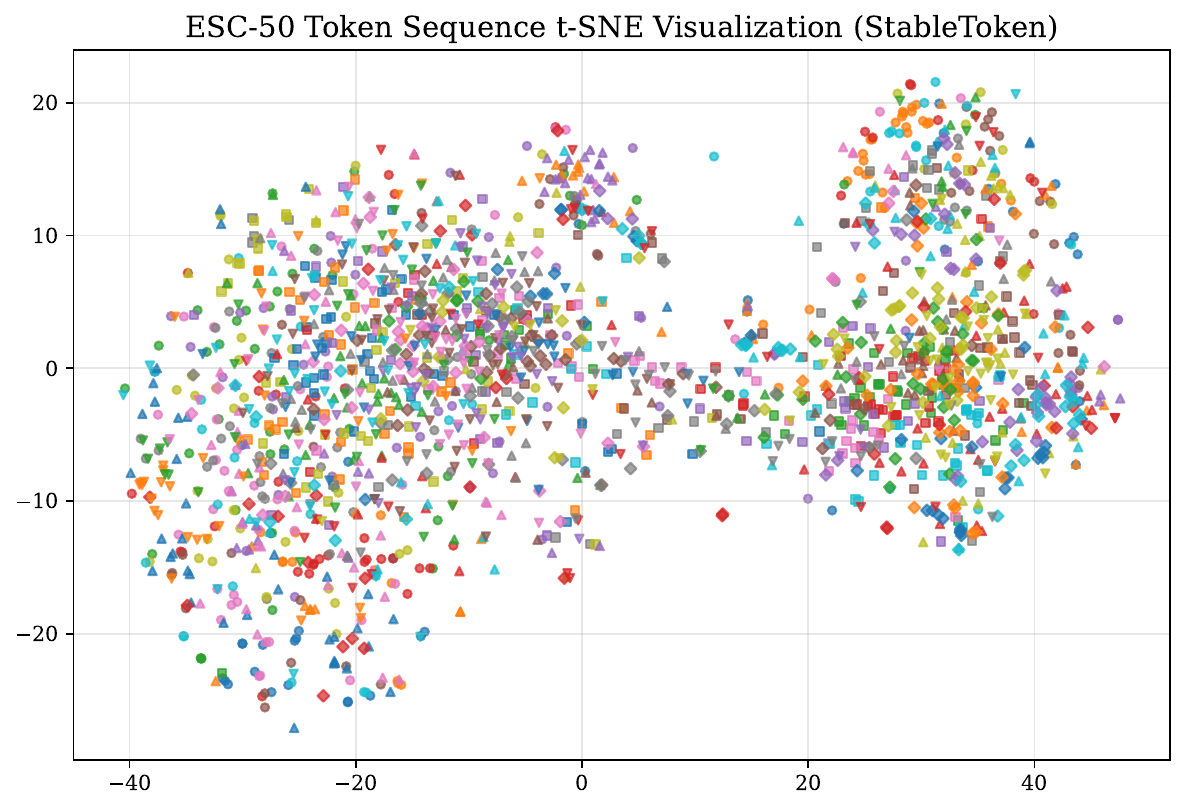}
        \caption{StableToken}
        \label{fig:tsne_sta}
    \end{subfigure}
    \hfill
    \begin{subfigure}{0.326666667\textwidth}
        \centering
        \includegraphics[width=\linewidth]{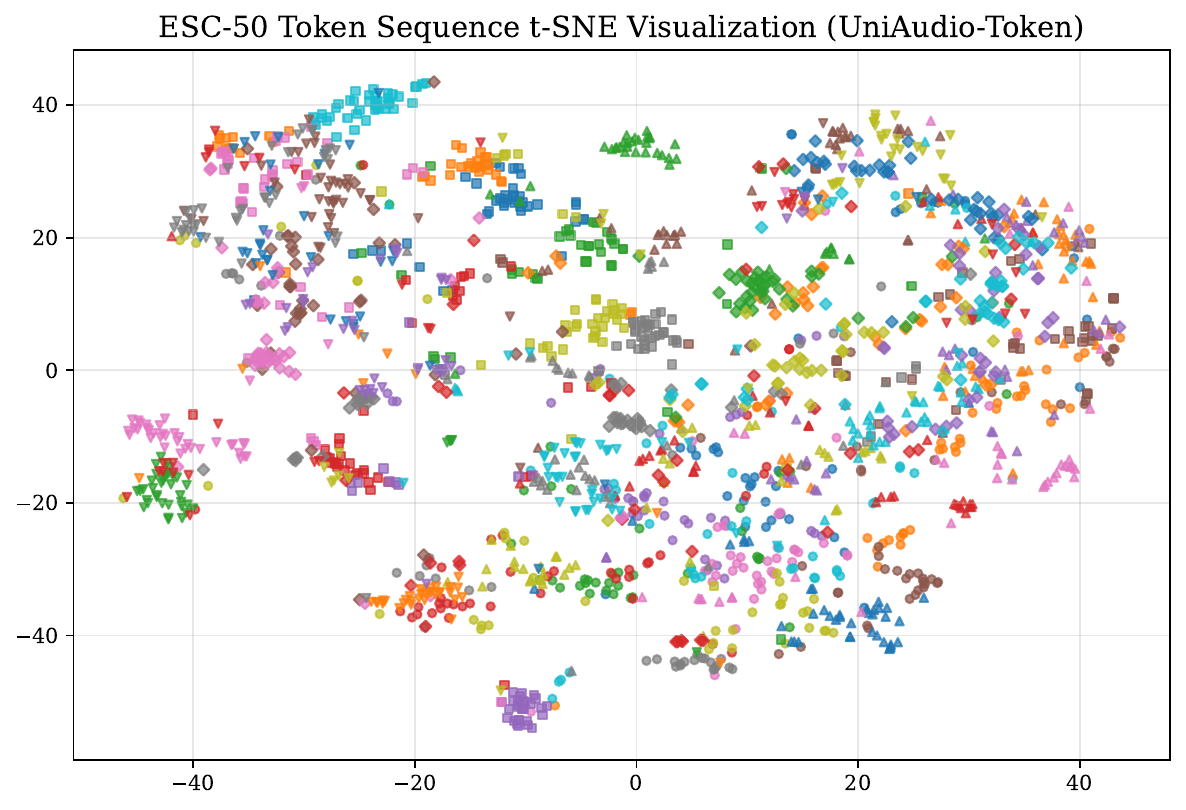}
        \caption{UniAudio-Token (Ours)}
        \label{fig:tsne_uni}
    \end{subfigure}
    \hfill
    \begin{subfigure}{0.326666667\textwidth}
        \centering
        \raisebox{0.065\height}{\includegraphics[width=\linewidth]{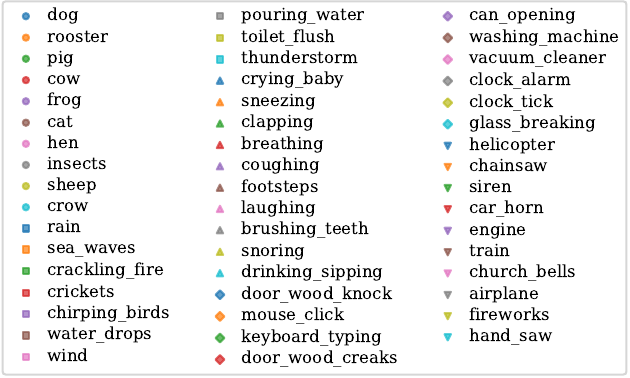}}
        \caption{Legend}
        \label{fig:esc50_legend}
    \end{subfigure}
    \caption{t-SNE visualization of token sequences on ESC-50. UniAudio-Token (Figure~\ref{fig:tsne_uni}) exhibits the most clear and semantically meaningful clusters, whereas the baselines show significant feature fragmentation and overlap.}
    \label{fig:esc50_tsne}
\end{figure*}

We evaluate UniAudio-Token from multiple perspectives. First, we examine its intrinsic quality through latent space analysis and speech reconstruction (\S\ref{sec:intrinsic}). Second, we assess its effectiveness as an interface for Audio-LLMs on both understanding and generation tasks (\S\ref{sec:extrinsic}). Finally, we analyze the contribution of SAE through ablation studies and mechanism visualizations (\S\ref{sec:ablation_analysis}).

%==============================================================================
\subsection{Tokenizer-Level Performance}
\label{sec:intrinsic}
%==============================================================================

\paragraph{Latent Space Disentanglement.}
To examine how discrete tokens capture granular acoustic characteristics, we visualize token sequences on ESC-50~\citep{piczak2015esc}, which is not included in our training data. 
ESC-50 covers a broad taxonomic range, from transient human sounds to stationary environmental textures and continuous mechanical noises. This comparison against baselines validates whether UniAudio-Token can perceive and disentangle diverse non-speech sounds.

Since standard dimensionality reduction techniques cannot be directly applied to discrete token sequences, we adopt a \textbf{Bag-of-Tokens} approach. For an audio clip with token sequence $T=  [t_0,$ $ t_1, \dots, t_n]$, we compute a token histogram vector:
\begin{equation}
    % \small
    H = [h_{0}, h_{1}, \dots, h_{V-1}]\in \mathbb{N}^{V},
\end{equation}
where $V$ is the codebook size, and the $i$-th element
\begin{equation}
    % \small
    h_{i} = \sum_{k=0}^n \mathbb{I}(t_k = i), \quad i=0,1,\dots,V-1,
\end{equation}
denotes the frequency of token ID $i$ in $T$. We subsequently apply t-SNE \cite{maaten2008visualizing} to project these high-dimensional histogram vectors into two dimensions for visualization.

Figure~\ref{fig:esc50_tsne} compares UniAudio-Token with baseline tokenizers. Baselines exhibit severe feature entanglement and fragmentation, while UniAudio-Token forms compact and well-separated clusters. This demonstrates that our method effectively captures acoustic characteristics of general audio, forming globally coherent and class-consistent representations.
Additional visualization results on the ESC-10 subset are provided in Appendix~\ref{app:esc10}.

\begin{table}[t]
\centering
\setlength{\tabcolsep}{4pt}
\resizebox{\linewidth}{!}{
\begin{tabular}{l cc cc}
\toprule
\multirow{2.5}{*}{\textbf{Model}} & \multicolumn{2}{c}{\textbf{ESC-10}} & \multicolumn{2}{c}{\textbf{ESC-50}} \\
\cmidrule(lr){2-3} \cmidrule(lr){4-5}
 & Sil. $\uparrow$ & Purity $\uparrow$ & Sil. $\uparrow$ & Purity $\uparrow$ \\
\midrule
WavTokenizer & -0.030 & 0.450 & -0.108 & 0.215 \\
GLM-4-Voice-Tokenizer & -0.182 & 0.373 & -0.304 & 0.133 \\
CosyVoice2 & -0.016 & 0.413 & -0.100 & 0.216 \\
StableToken & -0.035 & 0.468 & -0.096 & 0.174 \\
\midrule
\textbf{UniAudio-Token (Ours)} & \textbf{0.091} & \textbf{0.730} & \textbf{0.023} & \textbf{0.390} \\
\bottomrule
\end{tabular}
}
\caption{Clustering analysis on ESC-10 and ESC-50. UniAudio-Token is the only one achieving positive Silhouette Scores, indicating valid cluster separation.}
\label{tab:clustering_metrics_main}
\end{table}

To complement the qualitative visualization, we compute the Silhouette Score~\citep{rousseeuw1987silhouettes} and Cluster Purity~\citep{manning2008introduction} directly on the high-dimensional token histogram vectors to avoid information loss from dimensionality reduction. As shown in Table~\ref{tab:clustering_metrics_main}, UniAudio-Token is the only model achieving positive Silhouette Scores on both ESC-10 and ESC-50, while all baselines exhibit negative scores, indicating that their token distributions fail to form valid clusters aligned with acoustic categories. In terms of Cluster Purity, our model achieves huge improvements over baselines on both ESC-10 and ESC-50. These quantitative results support UniAudio-Token's meaningful discriminability across diverse acoustic events.

\begin{table*}[tb]
\centering
\resizebox{\textwidth}{!}{%
\begin{tabular}{l c c cccc c cccc c}
\toprule
& & & \multicolumn{5}{c}{\textbf{WER $\downarrow$}} & \multicolumn{5}{c}{\textbf{MOS $\uparrow$}} \\
\cmidrule(lr){4-8} \cmidrule(lr){9-13}
\textbf{Model} & \begin{tabular}[c]{@{}c@{}}\textbf{Frame}\\\textbf{Rate}\end{tabular} & \textbf{BPS} &
\begin{tabular}[c]{@{}c@{}}LS-\\clean\end{tabular} &
\begin{tabular}[c]{@{}c@{}}LS-\\other\end{tabular} &
\begin{tabular}[c]{@{}c@{}}SEED\\en\end{tabular} &
\begin{tabular}[c]{@{}c@{}}SEED\\zh\end{tabular} &
\textbf{Average} &
\begin{tabular}[c]{@{}c@{}}LS-\\clean\end{tabular} &
\begin{tabular}[c]{@{}c@{}}LS-\\other\end{tabular} &
\begin{tabular}[c]{@{}c@{}}SEED\\en\end{tabular} &
\begin{tabular}[c]{@{}c@{}}SEED\\zh\end{tabular} &
\textbf{Average} \\

\midrule
WavTokenizer & 75Hz & 900 & 5.07 & 13.09 & 5.60 & 4.02 & 6.95 & 3.37 & {3.09} & {3.01} & 3.13 & 3.15 \\
GLM-4-Voice-Tokenizer & 12.5Hz & 175 & 4.04 & 9.33 & 3.54 & 3.23 & 5.04 & 4.07 & {3.99} & \textbf{4.16} & 4.10 & 4.08 \\
CosyVoice2 & 25Hz & 325 & 4.25 & 9.68 & 4.34 & 2.75 & 5.26 & 3.36 & 3.25 & 3.31 & 3.58 & 3.38 \\
StableToken & 25Hz & 325 & {3.84} & {7.99} & {3.44} & {2.62} & {4.47} & {4.09} & 3.83 & 4.01 & 4.18 & {4.03} \\ \midrule
\textbf{UniAudio-Token (Ours)} & 25Hz & 325 & \textbf{3.47} & \textbf{6.79} & \textbf{2.55} & \textbf{1.90} & \textbf{3.68} \textcolor{blue}{\scriptsize{(-0.79)}} & \textbf{4.19} & \textbf{4.18} & 4.13 & \textbf{4.25} & \textbf{4.19} \textcolor{blue}{\scriptsize{(+0.11)}} \\
\bottomrule
\end{tabular}%
}
\caption{Speech reconstruction results measured via WER ($\downarrow$) and MOS ($\uparrow$).}
\label{tab:reconstruction}
\end{table*}

\begin{table*}[tb]
\centering
\resizebox{\textwidth}{!}{
\begin{tabular}{l cccc cccc ccc}
\toprule
\multirow{2}{*}{\textbf{Tokenizer}} & \multicolumn{4}{c}{\textbf{MMAU}} & \multicolumn{4}{c}{\textbf{MMAR}} & \multicolumn{3}{c}{\textbf{MMSU}} \\
\cmidrule(lr){2-5} \cmidrule(lr){6-9} \cmidrule(lr){10-12}
 & \textbf{Speech} & \textbf{Sound} & \textbf{Music} & \textbf{Overall} & \textbf{Speech} & \textbf{Sound} & \textbf{Music} & \textbf{Overall} & \textbf{Perception} & \textbf{Reasoning} & \textbf{Overall} \\ \midrule
WavTokenizer & 36.94 & 60.36 & 57.78 & 51.70 & 39.80 & 31.52 & 29.61 & 36.30 & 32.83 & 45.37 & 38.90 \\
CosyVoice2 & 39.94 & 61.56 & 62.57 & 54.70 & 41.50 & 35.76 & 30.58 & 38.10 & 27.44 & 45.83 & 36.34 \\
GLM-4-Voice-Tokenizer & 43.24 & 60.06 & 62.28 & 55.20 & 39.46 & 40.00 & 36.89 & 40.10 & 32.40 & 47.64 & 39.78 \\
StableToken & \textbf{45.05} & 58.56 & 55.99 & 53.20 & 42.18 & 39.39 & 31.07 & 39.10 & 31.98 & 49.71 & 40.56 \\ \midrule
\textbf{UniAudio-Token (Ours)} & \textbf{45.05} & \textbf{70.27} & \textbf{67.96} & \textbf{61.10} \textcolor{blue}{\scriptsize{(+5.90)}} & \textbf{45.24} & \textbf{43.64} & \textbf{40.29} & \textbf{45.80} \textcolor{blue}{\scriptsize{(+5.70)}} & \textbf{35.54} & \textbf{52.07} & \textbf{43.54} \textcolor{blue}{\scriptsize{(+2.98)}} \\ \bottomrule
\end{tabular}
}
\caption{Downstream Audio-LLMs audio understanding performance comparison, measured via accuracy (\%).}
\label{tab:main_results}
\end{table*}

% \begin{table}[htb]
% \centering
% \resizebox{\linewidth}{!}{%
% \begin{tabular}{l ccc}
% \toprule
% % \textbf{Tokenizer} & \textbf{LLM Base} &
% \textbf{Tokenizer} &
% SIM $\uparrow$ &
% WER $\downarrow$ &
% MOS $\uparrow$ &
% \midrule
% % CosyVoice2 & Qwen2.5-0.5B &
% CosyVoice2 &
% \textbf{.806} | .736  &
% 2.57 | 1.45 &
% 3.75 | 3.37 \\
% % \textbf{UniAudio-Token (Ours)} & Qwen2.5-0.5B &
% \textbf{UniAudio-Token (Ours)} &
% .792 | \textbf{.742} &
% \textbf{1.78} | \textbf{1.29} &
% \textbf{4.07} | \textbf{3.68} \\
% \bottomrule
% \end{tabular}%
% }
% \caption{TTS results measured via SIM ($\uparrow$), WER ($\downarrow$), and MOS ($\uparrow$) on SEED-TTS benchmark (en | zh).}
% \label{tab:tts}
% \end{table}

\paragraph{Speech Reconstruction Fidelity.}
% A key concern is whether the expansion to general audio would degrade performance on the core speech modality. To verify that our approach preserves speech reconstruction fidelity, we follow previous work~\cite{du2024cosyvoice2,zeng2024glm,song2025stabletoken} to train a flow matching model to reconstruct speech from discrete tokens, and assess performance via WER and MOS on LibriSpeech~\citep{panayotov2015librispeech} and SEED~\citep{anastassiou2024seed} benchmarks, where MOS is predicted by MOSNet~\citep{lo2019mosnet}. Results are presented in Table~\ref{tab:reconstruction}.

A universal tokenizer should improve general audio understanding without sacrificing speech generation. Following prior work~\cite{du2024cosyvoice2,zeng2024glm,song2025stabletoken}, we train a flow matching model to reconstruct speech from discrete tokens. We evaluate on LibriSpeech~\citep{panayotov2015librispeech} and SEED~\citep{anastassiou2024seed}, using WER and MOS (predicted by MOSNet~\citep{lo2019mosnet}).

Table~\ref{tab:reconstruction} shows that UniAudio-Token not only preserves, but also \textbf{further improves} speech reconstruction fidelity. It achieves a significantly lower WER and the highest average MOS. This indicates that retaining fine-grained acoustic cues can instead improve speech reconstruction capability.

% Notably, our UniAudio-Token not only preserves speech reconstruction capability but actually \textbf{surpasses} speech-centric baselines. Despite being a universal tokenizer, it achieves a significantly lower WER compared to semantic-centric baselines like GLM-4-Voice-Tokenizer and StableToken. This challenges the intuition that ASR-based compression yields better intelligibility. 

% We attribute this superiority to two primary factors: (1) \textbf{Inseparability of Content and Timbre:} Physically, linguistic content and acoustic style are deeply entangled~\citep{fant1971acoustic, polyak2019tts}.
% Semantic-centric tokenizers often aggressively strip away acoustic details to isolate pure meaning, inadvertently damaging phoneme integrity such as aspiration in consonants. By preserving these fine-grained acoustic cues via the SAE mechanism, our model allows the Flow Matching decoder to reconstruct clearer phonemes that are easier for ASR systems to recognize. (2) \textbf{Preservation of Accent:} We observe that baseline models often suffer from \textit{accent normalization}, reconstructing generic pronunciations that cause ASR mismatches (e.g., recognized as British ``colour''/``centre'' instead of the original American ``color''/``center''). In contrast, by preserving accent-specific acoustic features, our tokenizer facilitates the reconstruction to align with the original dialect, thereby reducing spelling errors in transcription.

We attribute this improvement to two factors: (1) \textit{Linguistic content and vocal attributes are not fully separable}~\citep{fant1971acoustic,polyak2019tts}. Overly aggressive semantic compression may remove acoustic details that are important for phonetic realization, such as aspiration and consonant transitions. SAE helps retain such cues, enabling clearer phoneme reconstruction. (2) \textit{UniAudio-Token better preserves accent-specific features}. In our analysis, baselines tend to normalize pronunciation, which can introduce transcription mismatches (e.g., ``colour'' vs. ``color'', and ``centre'' vs. ``center'') after reconstruction. By retaining accent characteristics, UniAudio-Token better matches original speech and reduces recognition errors.

% In summary, both latent space analysis and speech reconstruction results demonstrate that UniAudio-Token achieves strong intrinsic quality as a universal audio tokenizer.

Overall, the latent-space and reconstruction results show that UniAudio-Token provides a discrete representation both discriminative for general audio events and faithful for speech reconstruction.

%==============================================================================
\subsection{Downstream Audio-LLM Performance}
\label{sec:extrinsic}
%==============================================================================

% In the era of Audio-LLMs, the gold standard for evaluating audio tokenizers has evolved beyond mere reconstruction fidelity to their effectiveness as frontends for LLMs. Adopting this paradigm~\cite{zeng2024glm,song2025stabletoken}, we verify the representational quality of UniAudio-Token by employing it as the input interface for a standard LLM (Qwen2.5-3B). We assess the system-level performance on the mainstream MMAU~\citep{sakshi2025mmau}, MMAR~\citep{ma2025mmar}, and MMSU~\citep{wang2025mmsu} benchmarks, with all models instruction-tuned on the same data.

% A suitable audio tokenizer for Audio-LLMs should provide effective discrete representations for downstream understanding and generation. Following this paradigm~\citep{zeng2024glm,song2025stabletoken}, we integrate each tokenizer with the same Qwen2.5-3B backbone and evaluate on MMAU~\citep{sakshi2025mmau}, MMAR~\citep{ma2025mmar}, and MMSU~\citep{wang2025mmsu}. All systems are tuned under the same training setting for fair comparison.
A suitable audio tokenizer for Audio-LLMs should provide effective discrete representations for downstream \textbf{understanding} and \textbf{generation}. Following this paradigm~\citep{song2025stabletoken, du2024cosyvoice2}, we integrate each tokenizer with the same Qwen2.5-3B backbone to evaluate understanding, and the same Qwen2.5-0.5B backbone for generation. All systems are tuned identically for fair comparison.

\paragraph{Universal Audio Understanding.}
% As demonstrated in Table~\ref{tab:main_results}, the Audio-LLM equipped with UniAudio-Token achieves a significant performance lead over those using baseline tokenizers. In the \textbf{Speech} category, the model using the acoustic-centric WavTokenizer shows inferior performance, while UniAudio-Token enables the LLM to perform on par with or slightly better than semantic-centric tokenizers, confirming that our focus on universal audio does not compromise linguistic information retention. The most notable gap is observed in the \textbf{Sound} and \textbf{Music} tasks. The superiority in these categories indicates that UniAudio-Token retains environmental and musical cues that baseline semantic tokenizers discard as noise. In contrast, our proposed SAP supervision and SAE mechanism successfully preserve these cues, thereby providing the LLM with the necessary acoustic context to answer reasoning questions about complex environmental scenes and musical structures.

As shown in Table~\ref{tab:main_results}, UniAudio-Token yields the best performance across all three benchmarks. On \textbf{speech} tasks, UniAudio-Token matches or outperforms semantic tokenizers, confirming that its universal design does not compromise linguistic information. The largest improvements appear in \textbf{sound} and \textbf{music} categories, where baseline semantic tokenizers are limited by acoustic blindness and acoustic-centric tokenizers lack sufficient semantic structure. In contrast, SAP supervision encourages the codebook to encode vocal attributes and auditory-scene cues, while SAE adaptively restores shallow acoustic details. Together, these components provide the LLM with richer evidence for reasoning over complex sound events and musical content.

\begin{table}[tb]
\centering
\resizebox{\linewidth}{!}{
\begin{tabular}{l ccc}
\toprule
% \textbf{Tokenizer} & \textbf{LLM Base} &
\textbf{Tokenizer} &
SIM $\uparrow$ &
WER $\downarrow$ &
MOS $\uparrow$ \\
\midrule
% CosyVoice2 & Qwen2.5-0.5B &
CosyVoice2 &
.758 | \textbf{.762} | .760 &
2.71 | 1.39 | 2.05 &
3.75 | 3.37 | 3.56 \\
% \textbf{UniAudio-Token (Ours)} & Qwen2.5-0.5B &
\textbf{UniAudio-Token} &
\textbf{.792} | {.742} | \textbf{.767} &
\textbf{1.78} | \textbf{1.29} | \textbf{1.54} &
\textbf{4.07} | \textbf{3.68} | \textbf{3.88} \\
\bottomrule
\end{tabular}
}
\caption{TTS results measured via SIM ($\uparrow$), WER ($\downarrow$), and MOS ($\uparrow$) on SEED-TTS benchmark (en | zh | avg.).}
\label{tab:tts}
\end{table}

\paragraph{Controllable TTS Synthesis.}
We further assess UniAudio-Token on controllable text-to-speech (TTS) synthesis tasks. As other tokenizers do not support speaker embedding conditioning, we compare with CosyVoice2 on SEED-TTS. We condition on CAM++~\citep{wang23campp} and use ERes2Net~\citep{chen23eres2net} for speaker similarity (SIM) evaluation to avoid bias. Table~\ref{tab:tts} shows that UniAudio-Token yields significantly better WER and MOS, with slightly higher average SIM. These results prove that UniAudio-Token is also effective for autoregressive LLM-based speech generation.

%==============================================================================
\subsection{Analysis of SAE}
\label{sec:ablation_analysis}

\begin{figure*}[t]
    \centering
    \begin{subfigure}{0.49\textwidth}
        \centering
        \includegraphics[width=\linewidth]{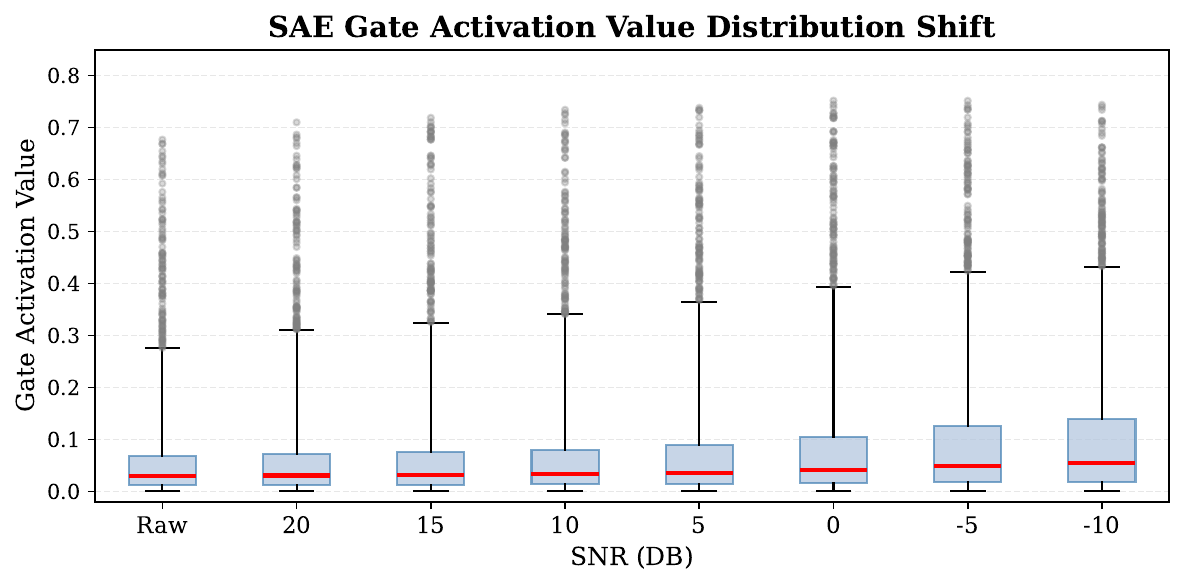}
        \caption{Noise-Adaptive Gating} % Leaves the (a) label but removes the text
        \label{fig:sae_stat}
    \end{subfigure}
    \hfill
    \begin{subfigure}{0.49\textwidth}
        \centering
        \includegraphics[width=\linewidth]{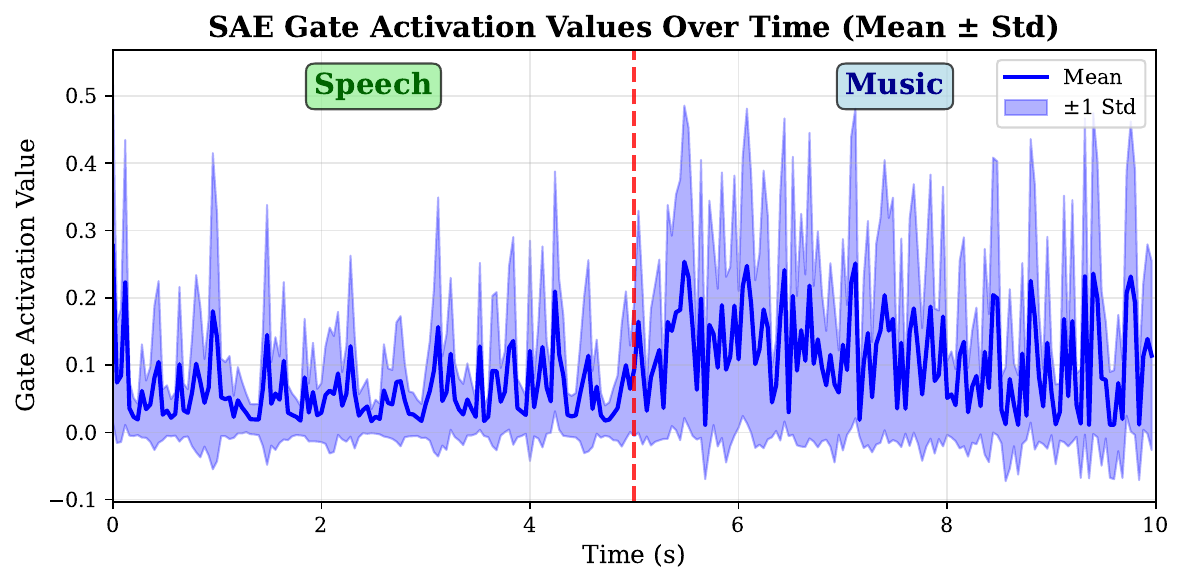}
        \caption{Modality-Aware Gating} % Leaves the (b) label but removes the text
        \label{fig:sae_case}
    \end{subfigure}
    \caption{Visualization of the SAE gate activation $\mathbf{g}$. The gate increases under lower SNR (Figure~\ref{fig:sae_stat}), and activates more strongly for music than speech (Figure~\ref{fig:sae_case}), demonstrating its content-aware dynamic behavior.}
    \label{fig:sae_analysis}
\end{figure*}
%==============================================================================

We next analyze SAE, a key mechanism of our framework. We first study the effect of fusion depth (\S\ref{sec:fusion_depth}), and then visualize the gate to verify its behavior as a content-aware adaptive mechanism rather than a static residual connection (\S\ref{sec:gating_behavior}).

\subsubsection{Impact of Fusion Depth}
\label{sec:fusion_depth}

% We first investigate the impact of injecting acoustic details from different encoder depths. We benchmark the model performance using two complementary metrics: (1) Word Error Rate (WER) on LibriSpeech~\citep{panayotov2015librispeech} to assess phonetic preservation, and (2) Non-Linguistic Score (NLS) on the AudioSet~\citep{audioset} to assess the ability to encode non-speech elements. A detailed NLS evaluation setup is provided in Appendix~\ref{app:nls}.

% We evaluate the effect of injecting acoustic features from different encoder layers. We use two complementary metrics: WER on LibriSpeech to assess phonetic preservation, and Non-Linguistic Score (NLS) on AudioSet~\citep{audioset} to measure acoustic information encoding. Details of NLS evaluation are provided in Appendix~\ref{app:nls}.
We evaluate the effect of injecting acoustic features from different encoder layers, using two complementary metrics: WER on LibriSpeech for phonetic preservation, and Non-Linguistic Score (NLS) on AudioSet~\citep{audioset} for acoustics. NLS evaluation details are provided in Appendix~\ref{app:nls}.
% ~\citep{panayotov2015librispeech}

\begin{table}[tb]
\centering
\resizebox{\linewidth}{!}{
\begin{tabular}{lccc}
\toprule
\multirow{2}{*}{\textbf{Configuration}} & \multicolumn{2}{c}{\textbf{WER (\%) $\downarrow$}} & \multirow{2}{*}{\textbf{NLS} $\uparrow$} \\
& LS-clean & LS-other & \\ \midrule
Baseline (w/o SAE)   & 2.47  & 5.71 & 2.93 \\
+ SAE ($L_1$)         & \textbf{2.41} & 5.62 & 3.08 \\
\textbf{+ SAE ($L_3$)} & 2.43 & \textbf{5.58} & \textbf{3.16} \\
+ SAE ($L_5$)         & 2.46 & 5.64 & 2.95 \\
\bottomrule
\end{tabular}
}
\caption{Impact of fusion depth in SAE. $L_k$ denotes using acoustic features from the $k$-th encoder layer.}
\label{tab:ablation}
\end{table}

As shown in Table~\ref{tab:ablation}, SAE consistently outperforms the baseline without SAE, especially on the challenging LS-other and NLS. This confirms that supplementing deep semantic features with shallow acoustic cues improves non-linguistic discriminability without harming phonetic preservation.

% Critically, the choice of fusion depth reveals a fundamental \textbf{semantic-acoustic trade-off}. Fusing from $L_1$ preserves the richest low-level waveform details (yielding the best WER on clean speech), but these features lack sufficient structural abstraction, creating a domain gap when fused with deep semantic representations. Conversely, $L_5$ features have undergone substantial semantic abstraction, filtering out most fine-grained acoustic cues, which explains its minimal improvement over the baseline on NLS (2.95 vs.\ 2.93).

The choice of fusion depth reveals a \textbf{semantic-acoustic trade-off}. $L_1$ features contain rich low-level details and yield the best WER on clean speech, but they are less structurally aligned with deep semantic representations. Conversely, $L_5$ features have undergone substantial semantic abstraction and therefore retain fewer fine-grained acoustic cues, resulting in only marginal NLS improvement.

% $L_3$ emerges as the \textbf{optimal equilibrium point}, achieving a peak NLS of 3.16 (8\% relative improvement over baseline) while maintaining competitive WER. This suggests that $L_3$ resides at a ``sweet spot'' in the encoder hierarchy---retaining sufficient acoustic cues (e.g., timbral textures, transient events) while being semantically abstract enough to harmonize with the deep features for effective fusion.

Fusion from $L_3$ provides the \textbf{optimal equilibrium}, achieving a peak NLS of 3.16 while maintaining competitive WER. This suggests that $L_3$ retains sufficient acoustic cues (e.g., timbral patterns and transient events) while remaining compatible with deep semantic features for effective fusion.

\subsubsection{Adaptive Gating Behavior}
\label{sec:gating_behavior}

% A natural question arises: does the learned gate $\mathbf{g}$ simply act as a static skip connection, or does it exhibit content-aware dynamic behavior? To verify that SAE functions as a genuinely adaptive mechanism, we visualize the gate activation under controlled acoustic conditions.

We further examine whether the learned gate acts as a content-aware adaptive controller. To this end, we statistically analyze and visualize the gate activations under controlled acoustic conditions.

\paragraph{Noise-Adaptive Gating.}
We mix clean speech in LibriSpeech with music from MusicBench~\citep{melechovsky2024mustango} at different Signal-to-Noise Ratios (SNRs). As shown in Figure~\ref{fig:sae_stat},
% there is a clear negative correlation between SNR and gate activation: as background music intensifies (lower SNR), the gate automatically opens wider to incorporate more low-level acoustic information. This noise-adaptive behavior explains our model's robustness on audio with complex acoustic backgrounds---the gate dynamically compensates for increased acoustic complexity by amplifying the contribution of shallow features.
gate activation increases as SNR decreases, indicating that when background noise (music) becomes more prominent, SAE injects more acoustic information to compensate for increased acoustic complexity.
% ~\citep{panayotov2015librispeech}

\paragraph{Modality-Aware Gating.}
We further examine the temporal dynamics of the gate on a concatenated clip containing 5 seconds of speech followed by 5 seconds of music. 
% Figure~\ref{fig:sae_case} reveals strikingly distinct activation patterns across modalities: during the speech segment, the gate remains largely suppressed, allowing the model to prioritize deep semantic abstraction for linguistic content; during the music segment, the gate exhibits high magnitude and variance, actively capturing the rich acoustic textures essential for non-linguistic audio.
Figure~\ref{fig:sae_case} reveals strikingly distinct activation patterns across modalities. During speech, the gate remains relatively suppressed, making the model rely more on deep semantic abstraction. During music, the gate becomes stronger and more variable, actively capturing acoustic textures essential for non-linguistic perception.

% These visualization results confirm our design intuition: the proposed SAE mechanism explicitly resolves the semantic-acoustic conflict by \textbf{decoupling the processing of semantic-dominant and acoustic-dominant signals}. Rather than applying a uniform fusion strategy, the gate learns to dynamically regulate information flow based on the instantaneous audio content, achieving a truly universal representation for both linguistic and non-linguistic audio.

These results support our motivation: rather than applying a fixed fusion strategy, SAE dynamically regulates the flow of shallow acoustic information according to the input content. This adaptive behavior helps UniAudio-Token balance linguistic abstraction with acoustic detail, yielding a unified representation for both speech and general audio.

\section{Conclusion}
\label{sec:conclusion}

% In this paper, we address the critical limitation of \textit{acoustic blindness} in current semantic speech tokenizers, which hinders their utility for general audio understanding. To bridge this gap, we introduce UniAudio-Token, a novel framework that empowers semantic tokenizers with comprehensive acoustic awareness without sacrificing their inherent linguistic strengths. By employing Semantic-Acoustic Primitives (SAP) as a structured supervision protocol and a Semantic-Acoustic Equilibrium (SAE) mechanism, we effectively rectified the information loss inherent in traditional semantic paradigms. Extensive experiments across diverse benchmarks demonstrate that UniAudio-Token consistently outperforms state-of-the-art baselines.

In this paper, we address the critical limitation of \textit{acoustic blindness} in current semantic speech tokenizers by introducing UniAudio-Token, a novel framework empowering semantic speech tokenizers with general audio perception. It leverages Semantic-Acoustic Primitives (SAP) as a supervision protocol and a Semantic-Acoustic Equilibrium (SAE) mechanism to adaptively rectify the acoustic information loss inherent in traditional semantic-centric paradigms. Extensive experiments validate its effectiveness in improving general audio discriminability, speech reconstruction fidelity, and downstream Audio-LLM performance.

\section*{Limitations}

% due to low BPS and text-based supervision, the reconstruction quality of complex non-speech audio naturally falls short of specialized high-bitrate acoustic codecs that are optimized for waveform reconstruction tasks. However, UniAudio-Token's reconstruction still retains identifiable temporal structures and rhythmic patterns of environmental sounds and music, providing the downstream LLM with acoustic cues for audio understanding.

UniAudio-Token is designed as a compact single-codebook tokenizer for Audio-LLMs, with an emphasis on balancing linguistic alignment, general audio perception, and speech generation. Due to its compact low-bitrate design and text-based supervision, its waveform-level reconstruction quality for complex non-speech audio still falls short of specialized high-bitrate acoustic codecs optimized for high-fidelity waveform reconstruction. 
% Nevertheless, UniAudio reconstructed signals retain identifiable temporal structures and rhythmic patterns, which provide useful acoustic cues for downstream LLMs. 
Moreover, while the UniAudio-Token framework itself is architecturally language-agnostic and can in principle support multilingual speech, our current training and evaluation mainly cover English and Chinese, constrained by the availability of audio-text data resources.
Future work may explore further improving non-speech audio reconstruction fidelity and more diverse language coverage, while preserving the compactness and semantic alignment required by LLM-based modeling.

% \section*{Acknowledgments}

% Bibliography entries for the entire Anthology, followed by custom entries
%\bibliography{anthology,custom}
% Custom bibliography entries only
\bibliography{references}

\appendix

\section{Samples of Semantic-Acoustic Primitives (SAP)}
\label{app:sap_sample}
In this section, we present some concrete examples of the Semantic-Acoustic Primitives (SAP) that are used as model supervision targets ($\mathbf{y}_{\text{SAP}}$) in our framework. Specifically, Figure~\ref{lst:sap_speech}, Figure~\ref{lst:sap_music}, and Figure~\ref{lst:sap_sound} illustrate examples of SAP data annotations associated with speech, music, and environmental sound audio clips, respectively.

\begin{figure*}[ht]
\begin{lstlisting}[language=json]
{
  %"linguistic_content"%: "I'd like to thank my husband for supporting me and going on this journey with me. Thank you.", 
  %"vocal_attributes"%: {
    #"age"#: "Adult",
    #"gender"#: "Female",
    #"emotion"#: "Happiness",
    #"accent"#: "General American English",
    #"prosody"#: "Slightly breathy, with a touch of vocal tremor and a natural rise and fall in pitch that conveys sincerity and emotion.",
    #"timbre"#: "Gentle"
  }, 
  %"auditory_scenes"%: {
    #"summary"#: "The audio captures a heartfelt personal acknowledgment in a large, reverberant hall, followed by a dense, energetic, and sustained wave of applause from a large audience, with natural reverberation matching the acoustics of the space.",
    #"events"#: [
      {
        #"class"#: "Audience applause",
        #"temporal_type"#: "impulsive",
        #"properties"#: "Dense, energetic, sustained"
      },
      {
        #"class"#: "Reverberation in hall",
        #"temporal_type"#: "persistent",
        #"properties"#: "Natural and sustained"
      }
    ]
  }
}
\end{lstlisting}
\caption{Example of SAP data annotation associated with a speech audio clip.}
\label{lst:sap_speech}
\end{figure*}

\paragraph{Speech.}
Figure~\ref{lst:sap_speech} illustrates the fine-grained annotation of a speech clip within a complex auditory environment. Beyond the verbatim transcript (the "linguistic\_content" field), the SAP schema also includes vocal attributes such as the speaker's identity and the expressive prosody. It also details a high-level audio clip summary alongside discrete auditory events, such as impulsive applause and hall reverberation. By encoding these as part of supervision targets, the framework facilitates the model to perceive the acoustic environment as a complement to speech components.

\begin{figure*}[htb]
\begin{lstlisting}[language=json]
{
  %"linguistic_content"%: null, 
  %"vocal_attributes"%: {
    #"age"#: null,
    #"gender"#: null,
    #"emotion"#: null,
    #"accent"#: null,
    #"prosody"#: null,
    #"timbre"#: null
  }, 
  %"auditory_scenes"%: {
    #"summary"#: "The audio clip is a high-fidelity, instrumental funk track characterized by a classic 1970s studio production style, with no vocals or extraneous sounds, and concludes with an abrupt, unresolved ending.",
    #"events"#: [
      {
        #"class"#: "Lead guitar riff",
        #"temporal_type"#: "persistent",
        #"properties"#: "Bright, slightly overdriven, rapid descending melodic phrase"
      },
      {
        #"class"#: "Rhythm guitar chords",
        #"temporal_type"#: "persistent",
        #"properties"#: "Sharp and staccato"
      },
      {
        #"class"#: "Drum pattern",
        #"temporal_type"#: "persistent",
        #"properties"#: "Punchy and syncopated"
      },
      {
        #"class"#: "Bass guitar line",
        #"temporal_type"#: "persistent",
        #"properties"#: "Melodic, syncopated, played with a pick"
      }
    ]
  }
}
\end{lstlisting}
\caption{Example of SAP data annotation associated with a music audio clip.}
\label{lst:sap_music}
\end{figure*}

\paragraph{Music.}
Figure~\ref{lst:sap_music} demonstrates an example of the SAP schema when encountering an instrumental music audio clip without a human voice. Due to the absence of speech, the ``linguistic\_content'' and ``vocal\_attributes'' fields are assigned \texttt{null} values. This sparsity prevents the model from hallucinating false linguistic and paralinguistic features in pure music audio clips. Moreover, this music instance prioritizes the ``events`` list to capture the concurrent instrumental layers. By separately supervising targets for each music component (such as lead guitar, rhythm guitar, drums, and bass), the framework encourages the model to learn disentangled representations of diverse instrument timbres and rhythmic patterns.

\begin{figure*}[htb]
\begin{lstlisting}[language=json]
{
  %"linguistic_content"%: null, 
  %"vocal_attributes"%: {
    #"age"#: null,
    #"gender"#: null,
    #"emotion"#: null,
    #"accent"#: null,
    #"prosody"#: null,
    #"timbre"#: null
  }, 
  %"auditory_scenes"%: {
    #"summary"#: "The audio is a low-fidelity field recording of a train passing through a tunnel or enclosed space, characterized by loud, rhythmic metallic clatter, deep rumble, and mechanical noise, followed by a sharp hiss from pneumatic brakes. The recording exhibits a reverberant, boomy acoustic environment, indicating an enclosed space. No human or animal presence is detected.",
    #"events"#: [
      {
        #"class"#: "Metallic clatter from train wheels",
        #"temporal_type"#: "impulsive",
        #"properties"#: "Loud, rhythmic"
      },
      {
        #"class"#: "Pneumatic brake hiss",
        #"temporal_type"#: "transient",
        #"properties"#: "Sharp, brief"
      },
      {
        #"class"#: "Train engine rumble",
        #"temporal_type"#: "persistent",
        #"properties"#: "Deep, continuous"
      },
      {
        #"class"#: "Background recording hiss",
        #"temporal_type"#: "persistent",
        #"properties"#: "faint, persistent"
      }
    ]
  }
}
\end{lstlisting}
\caption{Example of SAP data annotation associated with an environmental sounds audio clip.}
\label{lst:sap_sound}
\end{figure*}

\paragraph{Environmental Sounds.}
Figure~\ref{lst:sap_sound} shows the SAP's capability to categorize and supervise diverse acoustic phenomena in an environmental sound recording. It highlights the distinction between different temporal granularities, such as \textit{impulsive} wheel clatter, \textit{transient} brake hisses, and \textit{persistent} engine rumbles and background hiss. Therefore, the framework enables the model to learn representations that are sensitive to the start time and duration of acoustic events. Furthermore, the SAP captures not only the primary sound sources, but also the acoustic transformations (e.g., reverberation in Figure~\ref{lst:sap_speech}) imposed by the environment, thereby providing a foundation for downstream auditory reasoning tasks.

\section{Human Evaluation of SAP Data Annotation Quality}
\label{app:human_eval}

The reliability of the SAP data annotations is further verified through a manual quality assessment on a randomly sampled subset of the data by three human experts. Specifically, we conducted a manual audit on a uniformly random sample of 500 audio segments, covering speech, music, and environmental sounds. Three expert volunteers in audio processing performed the evaluation. The study adhered to ethical guidelines, with all participants providing informed consent and being notified of their right to withdraw without penalty. No financial incentives were involved in this process.

We validate the automated SAP generation by measuring the alignment between the audio content and the JSON outputs through the consensus of annotators. An attribute was considered valid only if at least two out of three experts confirmed its precision. Results are reported by the accuracy calculated over the eight fine-grained fields that comprise the vocal attributes and auditory scenes. We also calculated the 95\% Confidence Intervals (CI) using the Wilson score interval method.

% \paragraph{Evaluation Metric}
% For each sample, three annotators independently verified the alignment between the audio signal and the synthesized SAP JSON fields. An attribute was marked as "correct" only if the annotator consensus (majority vote) confirmed it accurately reflected the acoustic reality. We report the mean accuracy across the eight specific fields within the vocal attributes and auditory scenes domains.

% \begin{table}[htbp]
% \centering
% \caption{Human evaluation results of SAP data.}
% \label{tab:human_eval_results}
% % \resizebox{\linewidth}{!}{
% \begin{tabular}{llcc}
% \toprule
% \textbf{Domain} & \textbf{Field} & \textbf{Accuracy (\%)} & \textbf{95\% CI} \\ \midrule
% \multirow{6}{*}{Vocal Attributes} & Age & 98.8 & $[96.5, \, 99.6]$ \\
%  & Gender & 95.2 & $[91.8, \, 97.2]$ \\
%  & Emotion & 89.2 & $[84.7, \, 92.5]$ \\
%  & Accent & 96.0 & $[92.8, \, 97.8]$ \\
%  & Prosody & 96.0 & $[92.8, \, 97.8]$ \\
%  & Timbre & 95.6 & $[92.3, \, 97.5]$ \\ \midrule
% \multirow{2}{*}{Auditory Scenes} & Summary & 96.4 & $[93.3, \, 98.1]$ \\
%  & Events & 93.2 & $[89.4, \, 95.7]$ \\ \bottomrule
% \end{tabular}
% % }
% \end{table}

\begin{figure}[t]
    \centering
    \includegraphics[width=\linewidth]{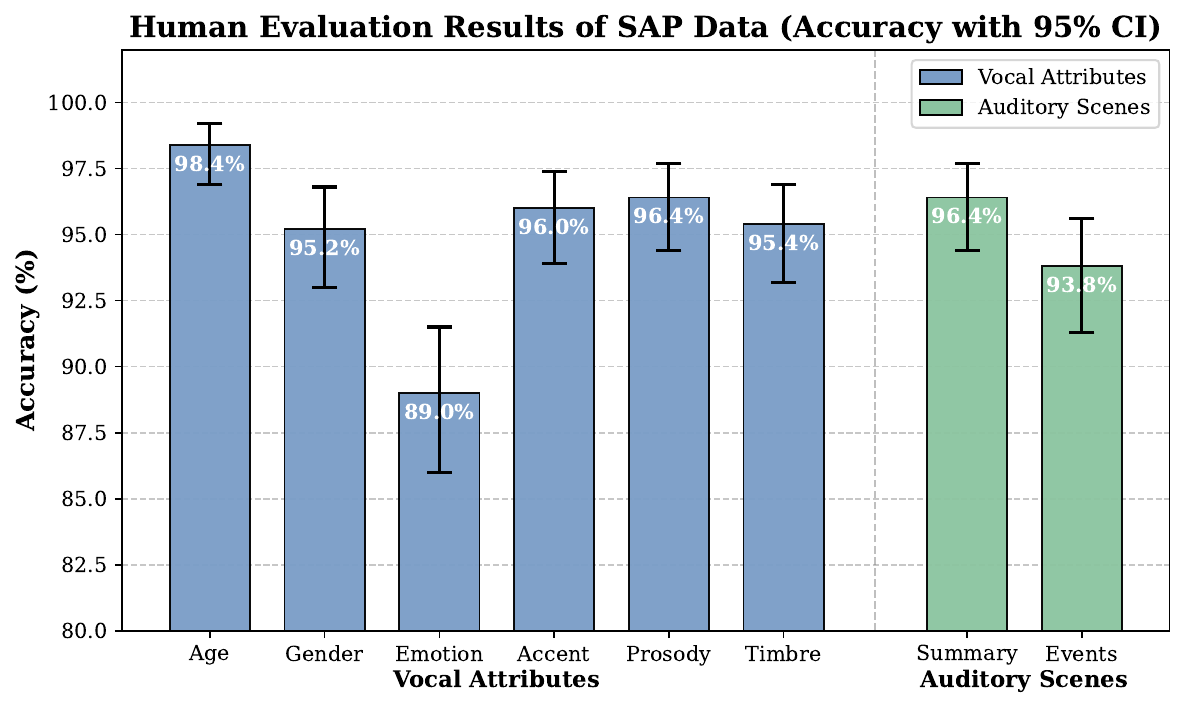}
    \caption{Human evaluation results of SAP data.}
    \label{fig:human_eval_results}
\end{figure}

% \paragraph{Quantitative Results}
% As shown in Table~\ref{tab:human_eval_results}, the SAP pipeline achieves high precision across most fields, with most paralinguistic and environmental attributes exceeding 95\% accuracy.

% \paragraph{Analysis} The evaluation reveals that stable biological and environmental traits (e.g., \textit{Age}, \textit{Gender}, and \textit{Continuous Events}) exhibit a high degree of reliability, with lower bounds of the 95\% confidence intervals consistently remaining above 92\%. This suggests that the pipeline effectively captures these relatively objective acoustic properties. The relative performance decrease in \textit{Emotion} (89.5\%, 95\% CI: [84.5, 93.1]) and \textit{Discrete Events} (90.0\%, 95\% CI: [85.1, 93.4]) reflects the inherent subjectivity of emotional state perception and the temporal sparsity of short-duration sound events. However, even accounting for statistical uncertainty, the accuracy for these challenging fields remains robustly above 84\%, demonstrating that the UAS pipeline provides a high-fidelity representation across all acoustic dimensions.

As detailed in Figure~\ref{fig:human_eval_results}, the automated SAP generation pipeline demonstrates great precision across the board, with the majority of attributes surpassing 95\% accuracy. It can be observed that objective vocal characteristics (e.g., \textit{Age} 98.4\%, \textit{Accent} 96.0\%) and high-level environmental descriptions (e.g., \textit{Summary} 96.4\%) exhibit high reliability, with the lower bounds of their 95\% confidence intervals consistently exceeding 92\%. The performance drop in \textit{Emotion} is likely attributable to the inherent subjectivity of perceiving emotional states, and the complexity of identifying all the transient and persistent acoustic cues without error or omission accounts for the lower accuracy in \textit{Events} list. Nonetheless, even for these nuanced fields, the lower confidence bound of accuracy remains above 85\%, confirming that the pipeline provides reliable supervision across vocal attributes and auditory scenes.

\section{Training Details of UniAudio-Token}
\label{app:training_details}

\subsection{Training Datasets for UniAudio-Token}
\label{app:datasets}

To develop a robust and versatile audio tokenizer, we trained UniAudio-Token on a massive, diverse corpus. Our training collection spans multiple domains, including high-quality speech, multi-lingual recordings, and diverse environmental sounds and music. Specifically, we incorporated major speech corpora such as Emilia (96,750 hours) and Yodas (29,155 hours) for broad linguistic coverage, alongside AudioSet (4,922 hours) to enhance the model's perception of non-linguistic acoustic events. A comprehensive summary of these open-source datasets and their respective durations is provided in Table~\ref{tab:uniaudio_token_datasets}.

% We train UniAudio-Token on hundreds of thousands of hours of both open-source data and in-house data. All open-source datasets used in this work are listed in Table~\ref{tab:uniaudio_token_datasets}.

\begin{table}[tb]
\centering
\resizebox{\linewidth}{!}{
\begin{tabular}{lr}
\toprule
\multirow{2}{*}{\textbf{Dataset}} & \textbf{Duration} \\
& \textbf{(\# Hours)} \\
\midrule
LibriSpeech~\citep{panayotov2015librispeech} & 960 \\
Multilingual LibriSpeech~\citep{pratap2020mls} & 27,322 \\
GigaSpeech~\citep{chen2021gigaspeech} & 10,000 \\
Yodas~\citep{li2023yodas} & 29,155 \\
Hi-Fi TTS~\citep{bakhturina2021hi} & 292 \\ 
VCTK~\citep{Veaux2017CSTRVC} & 44 \\
LibriTTS~\citep{zen2019libritts} & 586 \\
AISHELL-1~\citep{bu2017aishell} & 150  \\
WenetSpeech~\citep{zhang2022wenetspeech} & 10,005  \\
Common Voice~\citep{ardila2019common} & 2,133 \\
Emilia~\citep{he2024emilia} & 96,750 \\
AudioSet~\citep{audioset} & 4,922 \\
\bottomrule
\end{tabular}
}
\caption{Overview of public datasets included in the training of UniAudio-Token.}
\label{tab:uniaudio_token_datasets}
\end{table}

\subsection{Training Hyperparameters of UniAudio-Token}
\label{app:hyperparameters}

Table~\ref{tab:hyperparameters} summarizes the detailed hyperparameter configurations used during the training process of UniAudio-Token. To achieve a stable yet efficient optimization, we implement a multi-level learning rate strategy tailored to the specific functional roles of different model modules. Specifically, we apply a \textbf{conservative learning rate} ($1\text{e-}5$) to the pre-trained audio Encoder. This cautious approach is critical for preserving the rich semantic features and universal representations acquired during the large-scale foundational pre-training phase, thereby avoiding catastrophic forgetting. In contrast, the \textbf{Decoder utilizes a significantly higher learning rate} ($6\text{e-}4$) to facilitate the rapid acquisition of the novel Semantic-Acoustic Primitives (SAP) generation task. Other training configurations include:

\begin{itemize}
\item \textbf{Learning Rate Schedule}: A Cosine schedule with 1,000 warmup iterations is adopted to stabilize the initial gradient updates and provide a smooth transition towards convergence.

\item \textbf{Optimization Strategy}: We employ the AdamW optimizer ($\beta_1=0.9, \beta_2=0.999$) with a weight decay of $1\times10^{-2}$ to ensure robust generalization. \textbf{Gradient clipping} with a threshold of 1.0 is applied to prevent gradient explosion during training.

\item \textbf{Loss Balancing}: To maintain a high-quality discrete codebook while ensuring reconstruction fidelity, we set the \textbf{Quantization Loss Weight ($\lambda_1$)} to 10.0 and the \textbf{Commitment Loss Weight ($\lambda_2$)} to 2.5.
\end{itemize}

% \begin{table}[tb]
% \centering
% \resizebox{\linewidth}{!}{
% \begin{tabular}{ll}
% \toprule
% \textbf{Hyperparameter} & \textbf{Value} \\
% \midrule
% \multirow{3}{}{\text{Max Learning Rate}} & \multicolumn{1}{l}{Encoder: $1\text{e-}5$} \\
% & \multicolumn{1}{l}{Decoder: $6\text{e-}4$} \\
% & \multicolumn{1}{l}{Other: $2\text{e-}4$} \\
% % Encoder Max Learning Rate & $1\text{e-}5$ \\
% % Decoder Max Learning Rate & $6\text{e-}4$ \\
% % Other Parameters Max Learning Rate & $2\text{e-}4$ \\
% LR Schedule & \multicolumn{1}{l}{Cosine (with Linear Warmup)} \\
% Warmup Iterations & 1,000 \\
% Optimizer & \multicolumn{1}{l}{AdamW $\left(\beta_1=0.9, \beta_2=0.999\right)$} \\
% Weight Decay & \multicolumn{1}{l}{$1\text{e-}2$} \\
% Gradient Clipping & \multicolumn{1}{l}{1.0} \\
% Quantization Loss Weight ($\lambda_1$) & 10.0 \\
% Commitment Loss Weight ($\lambda_2$) & 2.5 \\
% \bottomrule
% \end{tabular}
% }
% \caption{Hyperparameter configurations across both training stages of UniAudio-Token.}
% \label{tab:hyperparameters}
% \end{table}

\begin{table}[tb]
\centering
\resizebox{\linewidth}{!}{
\begin{tabular}{ll}
\toprule
\textbf{Hyperparameter} & \textbf{Value} \\
\midrule
\multicolumn{2}{l}{\textit{\textbf{Learning rate}}} \\
\quad Encoder max LR & $1\times10^{-5}$ \\
\quad Decoder max LR & $6\times10^{-4}$ \\
\quad Other max LR & $2\times10^{-4}$ \\
LR schedule & Cosine (with linear warmup) \\
Warmup iterations & 1,000 \\
\midrule
\multicolumn{2}{l}{\textit{\textbf{Optimization}}} \\
Optimizer & AdamW $(\beta_1=0.9, \beta_2=0.999)$ \\
Weight decay & $1\times10^{-2}$ \\
Gradient clipping & 1.0 \\
\midrule
\multicolumn{2}{l}{\textit{\textbf{Loss weights}}} \\
Quantization loss $(\lambda_1)$ & 10.0 \\
Commitment loss $(\lambda_2)$ & 2.5 \\
\bottomrule
\end{tabular}
}
\caption{Hyperparameter configurations shared across the two training stages of UniAudio-Token.}
\label{tab:hyperparameters}
\end{table}

\section{Baseline Audio Tokenizers}
\label{app:baselines}
We compare UniAudio-Token against the following single-codebook audio tokenizers:
\begin{enumerate}
    \item \textbf{WavTokenizer}~\citep{ji2025wavtokenizer}, a high-compression single-codebook acoustic codecs. We use the officially released most powerful \textit{large, 75Hz} variant in our experiments.
    
    \item \textbf{CosyVoice2}~\citep{du2024cosyvoice2}, a leading speech tokenization and generation model, which introduces Finite-Scalar Quantization (FSQ) to replace traditional Vector Quantization (VQ) in its audio tokenizer for enhanced codebook utilization and representation efficiency.

    \item \textbf{GLM-4-Voice-Tokenizer}~\citep{zeng2025scaling}, a representative semantic tokenizer tailored for Speech Large Language Models. It can compress speech into highly efficient discrete tokens at a significantly lower frame rate while ensuring robust semantic preservation. We use the officially released checkpoint which has a frame rate of 12.5Hz and a codebook size of 16,384 in our experiments.

    \item \textbf{StableToken}~\citep{song2025stabletoken}, a novel semantic speech tokenizer with superior noise robustness. It employs a multi-branch Voting-LFQ architecture and adopts a bit-wise voting mechanism and a noise-aware training strategy to extract noise-irrelevant semantic speech tokens.
\end{enumerate}

\section{ESC-10 Token Sequence t-SNE Visualization Results}
\label{app:esc10}

\begin{figure*}[htb]
    \centering
    % 左图：UniAudio，右图：GLM-4
    \begin{subfigure}{0.326666667\textwidth}
        \centering
        \includegraphics[width=\linewidth]{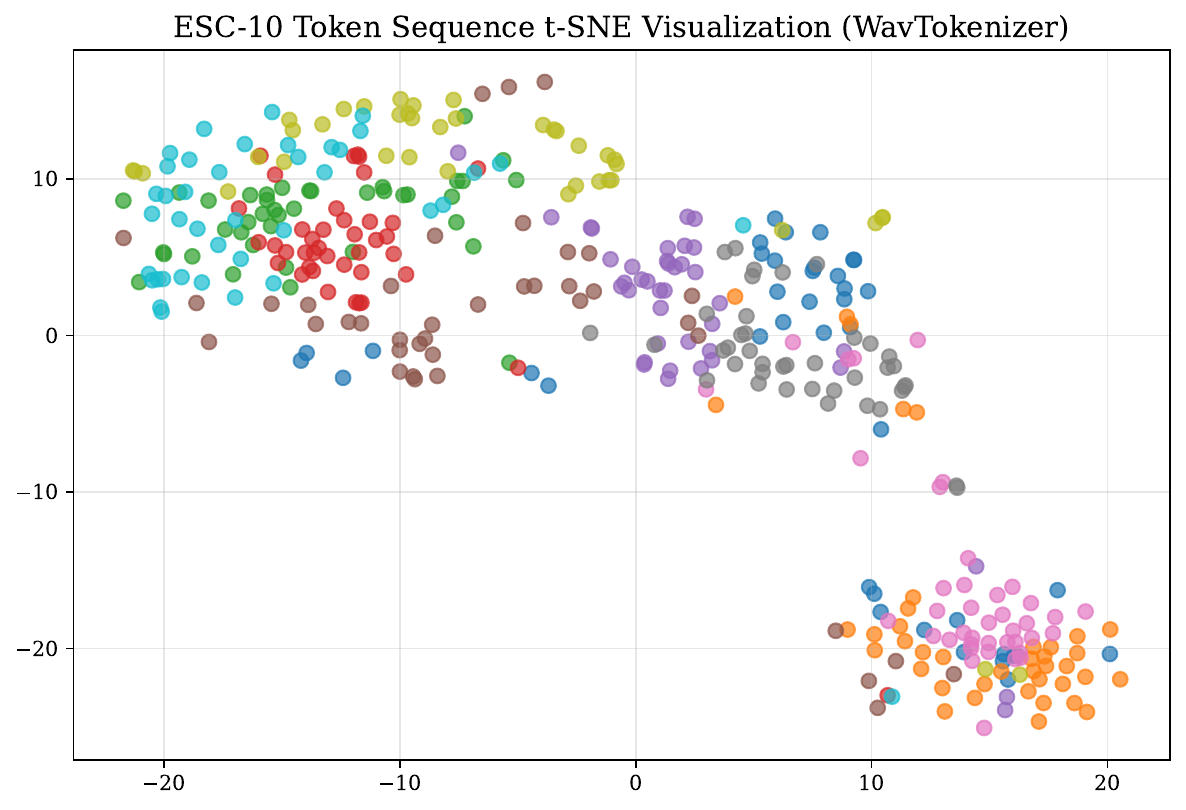}
        \caption{WavTokenizer}
        \label{fig:esc10_tsne_wav}
    \end{subfigure}
    % \hfill
    % \begin{subfigure}{0.495\textwidth}
    %     \centering
    %     \includegraphics[width=\linewidth]{pics/esc50/cos1_esc50.pdf}
    %     \caption{CosyVoice}
    %     \label{fig:esc50_tsne_cos1}
    % \end{subfigure}
    \hfill
    \begin{subfigure}{0.326666667\textwidth}
        \centering
        \includegraphics[width=\linewidth]{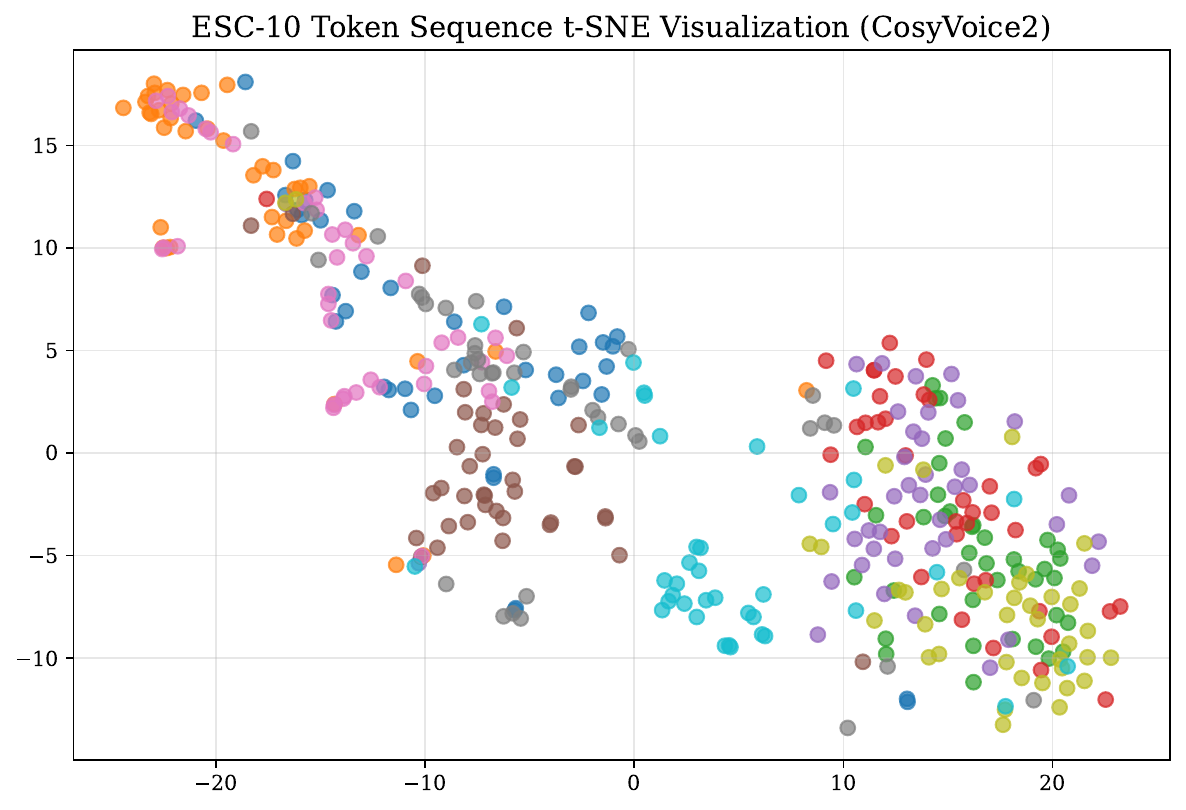}
        \caption{CosyVoice2}
        \label{fig:esc10_tsne_cos}
    \end{subfigure}
    \hfill
    \begin{subfigure}{0.326666667\textwidth}
        \centering
        \includegraphics[width=\linewidth]{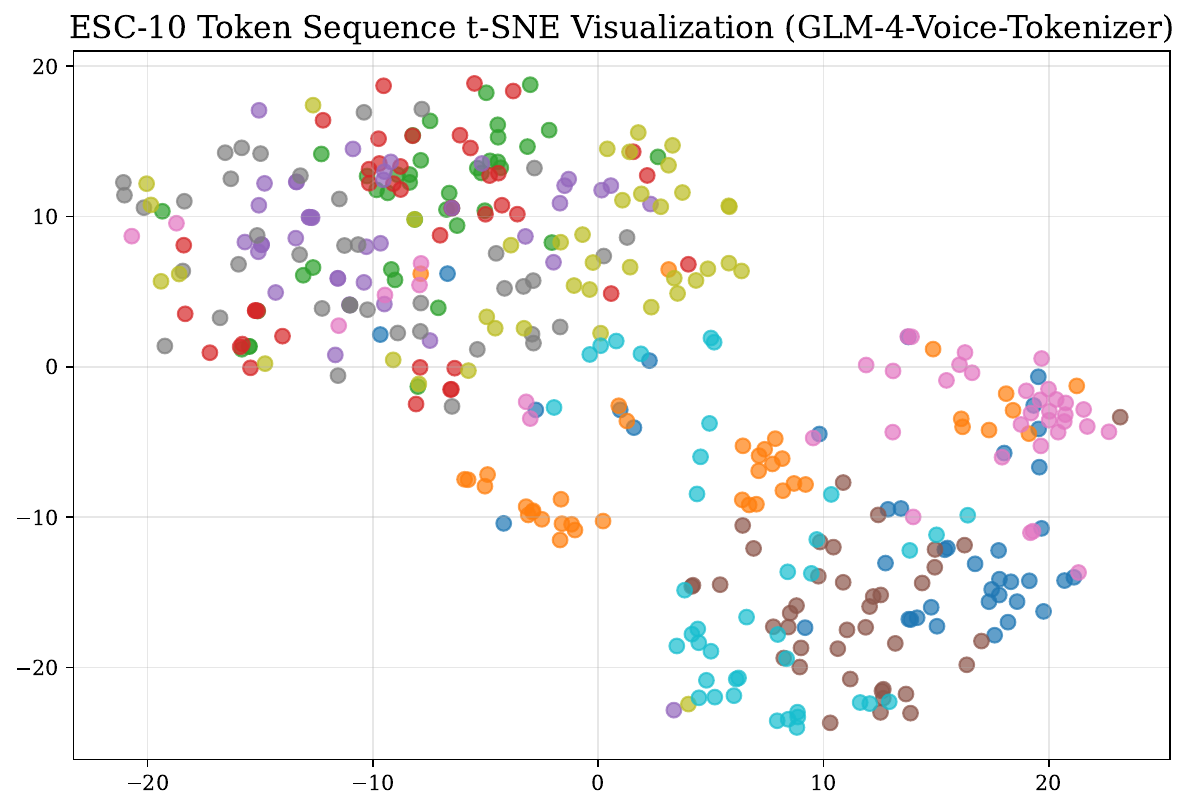} % 替换为你的文件名
        \caption{GLM-4-Voice-Tokenizer}
        \label{fig:esc10_tsne_glm}
    \end{subfigure}
    
    \begin{subfigure}{0.326666667\textwidth}
        \centering
        \includegraphics[width=\linewidth]{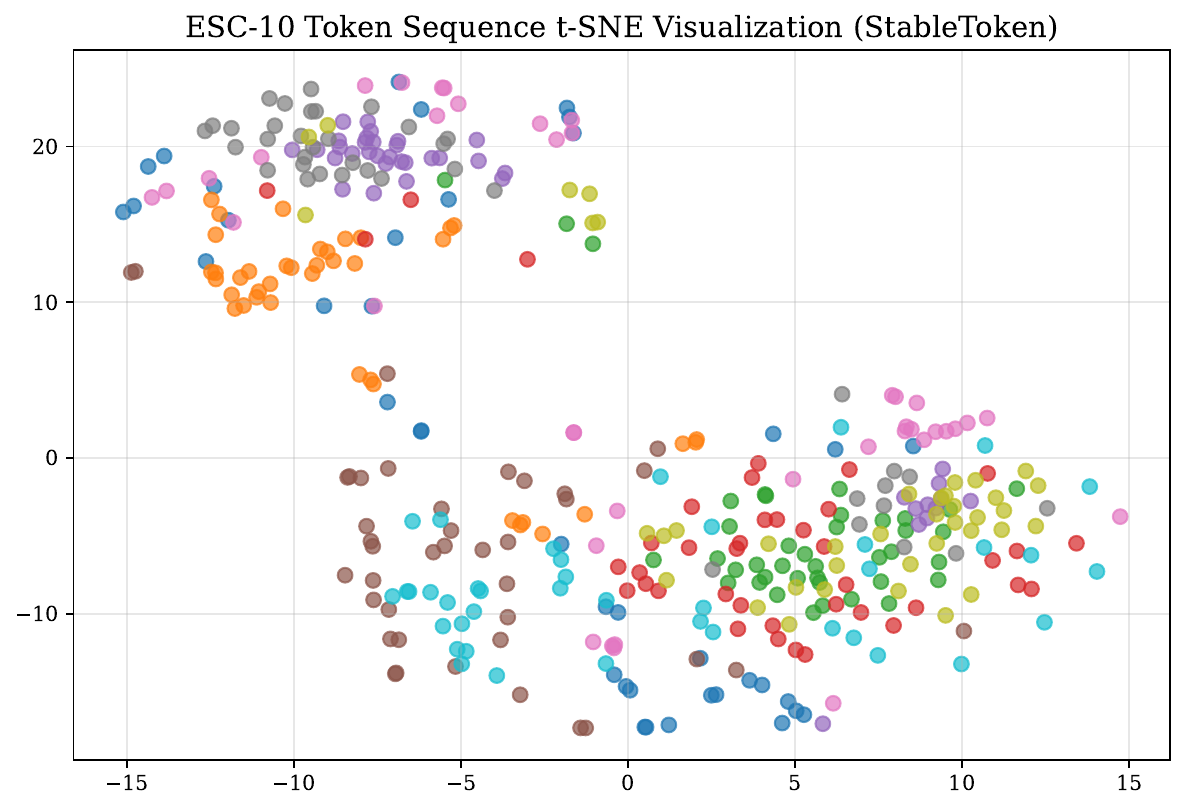}
        \caption{StableToken}
        \label{fig:esc10_tsne_sta}
    \end{subfigure}
    \hfill
    \begin{subfigure}{0.326666667\textwidth}
        \centering
        \includegraphics[width=\linewidth]{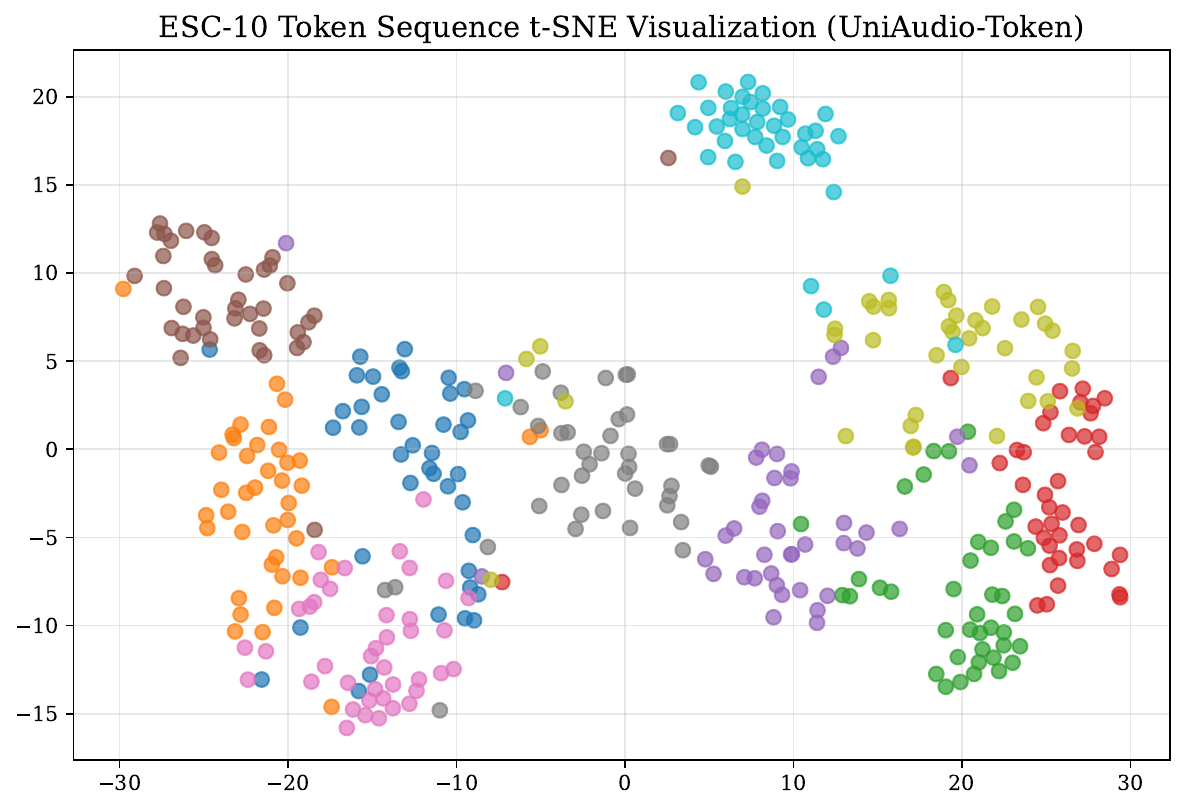} % 替换为你的文件名
        \caption{UniAudio-Token (Ours)}
        \label{fig:esc10_tsne_uni}
    \end{subfigure}
    \hfill
    \begin{subfigure}{0.326666667\textwidth}
        \centering
        \raisebox{0.53\height}{\includegraphics[width=0.9\linewidth]{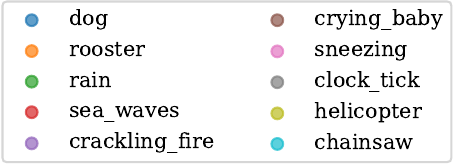}}
        \caption{Legend}
        \label{fig:esc10_legend}
    \end{subfigure}
    % \caption{t-SNE visualization of token histograms on the ESC-10 subset. (Left) \textbf{UniAudio-Token} forms tight, distinct clusters for each category, demonstrating robust fine-grained discrimination. (Right) \textbf{GLM-4-Voice-Tokenizer} suffers from severe fragmentation, particularly for environmental textures like rain and mechanical sounds, which are dispersed across the latent space.}
    \caption{t-SNE visualization of token histograms on the ESC-10 dataset. Our UniAudio-Token (Figure~\ref{fig:esc10_tsne_uni}) exhibits the most clear and semantically meaningful clusters, whereas the baselines (Figure~\ref{fig:esc10_tsne_wav},~\ref{fig:esc10_tsne_cos},~\ref{fig:esc10_tsne_glm},~\ref{fig:esc10_tsne_sta}) show significant feature fragmentation and overlap.}
    \label{fig:esc10_tsne}
\end{figure*}

To further investigate the semantic representation capabilities of different audio tokenizers, we provide a t-SNE visualization of the token histograms on the ESC-10 dataset in Figure \ref{fig:esc10_tsne}. This visualization maps high-dimensional token distributions into a 2D space to illustrate how well each model captures the underlying category information.

As shown in the figure, our UniAudio-Token (Figure \ref{fig:esc10_tsne}e) exhibits the most distinct and semantically meaningful clusters. Samples belonging to the same sound category (e.g., \textit{chainsaw}, \textit{rooster}, or \textit{sea waves}) are tightly grouped together with clear boundaries. In contrast, the baseline models, including WavTokenizer, CosyVoice2, GLM-4-Voice-Tokenizer, and StableToken (Figures~\ref{fig:esc10_tsne_wav}--\ref{fig:esc10_tsne_sta}), show significant feature fragmentation and overlap. In these baselines, different classes are often intermingled, suggesting that their token sequences lack sufficient discriminative power for environmental sound classification. These results demonstrate that UniAudio-Token's discrete representations are more effective at capturing high-level semantic features compared to existing speech-centric or general-purpose audio tokenizers.

\section{Non-Linguistic Score Evaluation Setup}
\label{app:nls}
In this section, we detail the methodology for calculating the Non-Linguistic Score (NLS), a metric designed to quantify the truthfulness of model-generated audio descriptions. Unlike traditional n-gram based metrics (e.g., BLEU or METEOR), the NLS focuses on the high-level alignment between generated content and the annotations provided in the SAP dataset.

To ensure a robust and nuanced evaluation, we employ an LLM-based judge framework. Specifically, we utilize the {Qwen3-235B-A22B-Instruct-2507}~\citep{yang2025qwen3} model to perform zero-shot scoring. This approach leverages the model's advanced reasoning capabilities to move beyond simple keyword matching, allowing for an assessment of complex audio attributes such as sound event sources, environmental context (e.g., spatial hints or reverberation), and audio characteristics (e.g., fidelity and mix).

The evaluation process is standardized through a carefully constructed prompt, as illustrated in Figure \ref{fig:nls_prompt}. The judge is tasked with assigning a score on a scale from 1 to 5, where a score of 5 represents "Perfect Consistency" and a score of 1 indicates a fundamental divergence in content. To minimize variance and maintain consistency across evaluations, the prompt provides explicit scoring criteria for each level, focusing on the preservation of principal information and the presence of contradictions. By enforcing a strict output format (score only), we ensure the results are directly parsable for large-scale quantitative analysis.

\begin{figure*}[!htbp]
    % 使用我们定义的 promptbox
    \begin{promptbox2}{Prompt for Evaluating Non-Linguistic Score (NLS)}
You are given two descriptions of the same audio clip:

- **Reference Description:** The ground-truth expert annotation describing the audio in detail.
- **Generated Description:** The generated description produced by a model.

Your task is to judge how well the Generated Description matches the Reference Description in its description of the audio clip.  
**Audio can be any type**-music, singing, speech, spoken dialogue, environmental sounds, animal noises, or other non-musical audio.

**Please consider these aspects as relevant (if present):
- **Sound events & sources:** Describe what sounds are present (e.g. music, speech, environmental sounds, specific objects, animals, vehicles, etc.)
- **Environmental context:** Indoors/outdoors, location hints, crowd noise, weather, reverberation, etc.  
- **Audio characteristics:** Quality, fidelity, mix, spatial placement, background/foreground, clarity or distortion  

**Scoring Criteria:**  
Assign a score from **1** to **5** based on these standards:

- **5 - Perfect Consistency:**  
  The Generated Description accurately covers almost all important information given in the Reference Description, including types of sound, main events, vocal or non-vocal details, environmental context, and tone. Any differences are minor and inconsequential.

- **4 - Mostly Consistent:**  
  The Generated Description includes most major elements from the Reference Description. Some small details or less important aspects may be missing or stated differently, but all principal information matches closely. Only minor mistakes or omissions are present.

- **3 - Partially Consistent:**  
  The Generated Description captures some main aspects correctly, but omits important details or contains noticeable errors regarding sound types, vocal or environmental features, or context. There is reasonable overlap, but inconsistency in parts.

- **2 - Mostly Inconsistent:**  
  The Generated Description reflects very few features from the Reference Description. Most elements are missing, inaccurate, or contradicted (e.g., confuses speech for music; wrong language; misses major sound sources; incorrect context). There is substantial disagreement.

- **1 - Completely Inconsistent:**  
  The Generated Description does not match the Reference Description in content or meaning. Major sound events/types, vocal or environmental information, and context differ fundamentally.

Below are the Generated Description and the Reference Description:

### [Generated Description]: ${generated}

### [Reference Description]: ${reference}

After evaluating, please output the score only without anything else. You don't need to provide any explanations.

    \end{promptbox2}
\caption{Prompt template used by Qwen3-235B-A22B-Instruct-2507 for Non-Linguistic Score (NLS) evaluation.}
\label{fig:nls_prompt}
\end{figure*}

\section{LLM Usage Statement}

In accordance with ACL policy on Generative AI tools and technologies, the authors hereby disclose the following: After the authors completed the initial draft of this paper, LLMs were utilized to enhance grammar and polish the writing of this manuscript. No new research ideas, experimental designs, or scientific content were generated by the LLMs. All scientific contributions, analyses, and conclusions presented in this work are solely those of the authors. We take full responsibility for the content of this paper, including all sections that have been revised or improved with LLM assistance. The LLMs are not authors and did not contribute to the research ideation or substantive scientific writing.

This statement is provided to ensure transparency and compliance with the ACL Guidelines on Generative Assistance in Authorship.

\end{document}